\documentclass[letterpaper]{article} 
\usepackage[submission]{aaai25}  
\usepackage{times}  
\usepackage{helvet}  
\usepackage{courier}  
\usepackage[hyphens]{url}  
\usepackage{graphicx} 
\usepackage{natbib}  
\usepackage{caption} 
\usepackage{algorithm}
\usepackage{algorithmic}

\usepackage{graphicx}
\usepackage{amsmath}
\usepackage{amssymb}
\usepackage{booktabs}
\usepackage{pifont}
\usepackage{times}
\usepackage{soul}
\usepackage{url}
\usepackage{multirow}
\usepackage[utf8]{inputenc}
\usepackage{wrapfig}
\usepackage{tcolorbox}
\usepackage{listings}
\usepackage{framed}
\usepackage{textcomp}
\usepackage{minitoc}
\usepackage{colortbl}
\usepackage{xcolor}
\usepackage{algorithm}
\usepackage{algorithmic}
\usepackage{amssymb}
\usepackage{xcolor}
\usepackage{pifont}
\usepackage{booktabs}  
\usepackage{colortbl}  
\usepackage{multirow}  
\usepackage{array}
\usepackage{graphicx}  
\usepackage{xcolor}    
\usepackage{pifont}    
\usepackage{tikz}
\usepackage{hyperref}
\usepackage{minitoc}

\usepackage[capitalize]{cleveref}
\crefname{section}{Sec.}{Secs.}
\Crefname{section}{Section}{Sections}
\Crefname{table}{Table}{Tables}
\crefname{table}{Tab.}{Tabs.}

\usepackage{newfloat}
\usepackage{listings}
\DeclareCaptionStyle{ruled}{labelfont=normalfont,labelsep=colon,strut=off} 
\floatstyle{ruled}
\newfloat{listing}{tb}{lst}{}
\floatname{listing}{Listing}

\title{%
  \makebox[0pt][l]{\raisebox{-0.4\height}{\includegraphics[height=1cm]{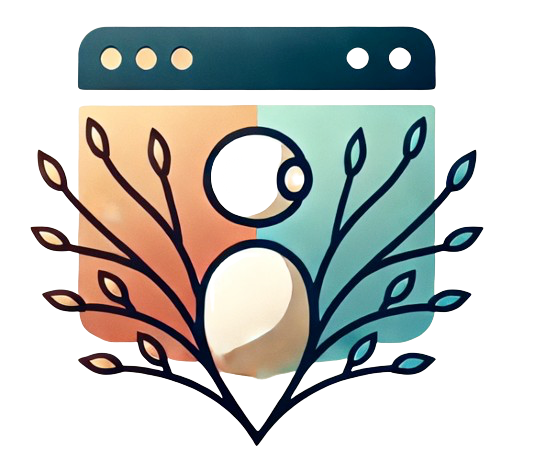}}}%
  \makebox[\dimexpr\textwidth+0.8cm\relax][c]{WebPilot: A Versatile and Autonomous Multi-Agent System for Web Task }\\
  \makebox[\dimexpr\textwidth+1cm\relax][c]{Execution with Strategic Exploration}
}

\author{Yao Zhang \textsuperscript{\rm 1,3}\thanks{Equal contribution} \thanks{Corresponding authors}
\qquad Zijian Ma \textsuperscript{\rm 2}\footnotemark[1] 
\qquad Yunpu Ma\textsuperscript{\rm 1,3}\footnotemark[2] \\
\qquad Zhen Han \textsuperscript{\rm 1}
\qquad Yu Wu \textsuperscript{\rm 2}
\qquad Volker Tresp\textsuperscript{\rm 1,3}\footnotemark[2]\\
    \textsuperscript{\rm 1} LMU Munich \qquad  
    \textsuperscript{\rm 2} Technical University of Munich 
    \textsuperscript{\rm 3} Munich Center for Machine Learning \\ 
    yzhang@dbs.ifi.lmu.de \qquad tresp@dbs.ifi.lmu.de
   	}

\usepackage{bibentry}

\begin{document}
\pagestyle{plain}

\maketitle

\begin{abstract}
LLM-based autonomous agents often fail to execute complex web tasks that require dynamic interaction, largely due to the inherent uncertainty and complexity of these environments. Existing LLM-based web agents typically rely on rigid, expert-designed policies specific to certain states and actions, lacking the flexibility and generalizability needed to adapt to unseen tasks. In contrast, humans excel by exploring unknowns, continuously adapting strategies based on new observations, and resolving ambiguities through exploration. To emulate human-like adaptability, web agents need strategic exploration and complex decision-making. Monte Carlo Tree Search (MCTS) is well-suited for this, but classical MCTS struggles with vast action spaces, unpredictable state transitions, and incomplete information in web tasks. In light of this, we develop WebPilot, a multi-agent system with a dual optimization strategy that improves MCTS to better handle complex web environments. Specifically, the Global Optimization phase involves generating a high-level plan by breaking down tasks into manageable subtasks, continuously refining this plan through reflective analysis of new observations and previous subtask attempts, thereby focusing the search process and mitigating challenges posed by vast action spaces in classical MCTS. Subsequently, the Local Optimization phase executes each subtask using a tailored MCTS designed for complex environments, effectively addressing uncertainties and managing incomplete information by iteratively refining decisions based on new observations. Experimental results on WebArena and MiniWoB++ demonstrate the effectiveness of WebPilot. Notably, on WebArena, WebPilot achieves SOTA performance with GPT-4, achieving a \textbf{93\% relative increase} in success rate over the concurrent tree search-based method. WebPilot marks a significant advancement in general autonomous agent capabilities, paving the way for more advanced and reliable decision-making in practical environments. Our code is publicly released at \href{https://yaoz720.github.io/WebPilot/}{github.com/WebPilot}.

\end{abstract}

\begin{table*}[h]
\centering
\scalebox{0.88}{
\begin{tabular}{l>{\centering\arraybackslash}m{1.8cm}
	>{\centering\arraybackslash}m{1.8cm}
	>{\centering\arraybackslash}m{1.8cm}
	>{\centering\arraybackslash}m{1.8cm}
	>{\centering\arraybackslash}m{1.8cm}
	>{\centering\arraybackslash}m{1.8cm}}
\toprule
\textbf{Benchmark} & \textbf{Dynamic Interaction} & \textbf{Partial Obs. Env.} & \textbf{Non-Fixed Policy} & \textbf{Scalable} & \textbf{Realistic Web Env.} & {\textbf{Comp. Self-Reward}} \\ 
\midrule
\textbf{RAP \cite{hao2023reasoning}} & \textcolor{red}{\ding{55}} & \textcolor{red}{\ding{55}} & \textcolor{red}{\ding{55}} & \textcolor{red}{\ding{55}} & \textcolor{red}{\ding{55}} & \textcolor{red}{\ding{55}}\\ 
\textbf{LATS \cite{zhou2023language}} & \textcolor{green}{\checkmark} & \textcolor{red}{\ding{55}} & \textcolor{red}{\ding{55}} & \textcolor{red}{\ding{55}} & \textcolor{red}{\ding{55}} & \textcolor{red}{\ding{55}}\\ 
\textbf{LLM-MCTS \cite{zhao2024large}} & \textcolor{red}{\ding{55}} & \textcolor{green}{\checkmark} & \textcolor{red}{\ding{55}} & \textcolor{green}{\checkmark} & \textcolor{red}{\ding{55}} &\textcolor{red}{\ding{55}}\\ 
\midrule
\textbf{SteP \cite{sodhi2024step}} & \textcolor{green}{\checkmark} & \textcolor{green}{\checkmark} & \textcolor{red}{\ding{55}} & \textcolor{red}{\ding{55}} & \textcolor{green}{\checkmark} & \textcolor{red}{\ding{55}}\\ 
\textbf{LM-Tree Search \cite{koh2024treeCMU}} & \textcolor{green}{\checkmark} & \textcolor{green}{\checkmark} & \textcolor{green}{\checkmark} & \textcolor{red}{\ding{55}} & \textcolor{green}{\checkmark} & \textcolor{red}{\ding{55}}\\ \hline
\textbf{WebPilot} & \textcolor{green}{\checkmark} & \textcolor{green}{\checkmark} & \textcolor{green}{\checkmark} & \textcolor{green}{\checkmark} & \textcolor{green}{\checkmark} & \textcolor{green}{\checkmark}\\ 
\bottomrule
\end{tabular}
}
\vspace{-5pt}
\caption{Comparison of different agent types, including web agents and MCTS-based agents. Partial Obs. Env. - Partially Observable Environment. Comp. Self-Reward - Comprehensive self-reward mechanism.}
\vspace{-10pt}
\label{table:benchmark-comparison}
\end{table*}

\section{Introduction}
The advanced reasoning capabilities of Large Language Models (LLMs) \cite{yang2023dawn, achiam2023gpt, team2023gemini, anthropic2024claude} have significantly expanded the potential for developing autonomous web agents capable of navigating and interacting within complex, dynamic environments \cite{lai2024autowebglm,
deng2024mind2web}. To fully harness this potential, these agents must excel in tasks such as complex information retrieval, long-horizon task execution, and the integration of diverse information sources \cite{wang2024survey, zhou2023webarena}.

However, despite the advanced reasoning capabilities of LLMs, current LLM-based web agents \cite{sodhi2024step,ma2023laser} often fall short in executing complex web tasks that require dynamic interaction. This limitation arises primarily from their heavy reliance on rigid, expert-designed policies tailored to specific states and actions. While these policies are meticulously crafted to address well-defined scenarios, they inherently lack the flexibility and generalizability needed to adapt to the uncertain and variable nature of real-world web environments, as well as to unseen tasks.

In contrast, humans excel at handling complex web tasks due to their cognitive flexibility \cite{daw2005uncertainty}, which allows them to explore unknowns, adjust plans dynamically based on new observations, and resolve ambiguities through trial and error. This adaptability enables humans to navigate uncertain environments, make decisions with incomplete information, and modify strategies in real time. Monte Carlo Tree Search (MCTS) mirrors this cognitive process, making it particularly effective in emulating human web navigation strategies. MCTS facilitates the exploration of unknowns by expanding nodes during the tree search, helping web agents discover effective actions. It adjusts tactical-level strategies during the search process, refining action generation at each node based on the feedback from the expansion, much like how humans iteratively adjust their actions in response to new observations. When encountering dead ends or unclear paths—nodes with low potential or uncertain outcomes—MCTS reassesses and explores alternative branches, effectively addressing the limitations of LLMs in handling unfamiliar web environments.

Despite its potential, classical MCTS \cite{browne2012survey} struggles in complex web environments due to vast action spaces, unpredictable state transitions, and incomplete information. While recent methods like LLM-MCTS \cite{zhao2024large} and LATS \cite{zhou2023language} integrate LLMs for heuristic guidance, they are limited to tasks with smaller action spaces and lower complexity, reducing their effectiveness in real-world scenarios. Similarly, RAP \cite{hao-etal-2023-reasoning} optimizes inference paths but lacks the flexibility for dynamic interaction in complex environments. Reward mechanism within MCTS also remains challenging for complex environments; current approaches either rely on direct environment reward \cite{zhou2023language}, which is impractical for real-world tasks, or use overly simplistic scoring systems (e.g., binary or low-resolution scales \cite{koh2024treeCMU}), failing to accurately capture the nuanced and evolving nature of web environments. These limitations highlight the need for more adaptable and robust MCTS-based methods capable of effectively navigating complex web tasks.

In response to these challenges, we introduce WebPilot, a versatile multi-agent system designed with a dual optimization strategy grounded in the principles of MCTS, specifically tailored for enhanced adaptability in complex environments. WebPilot first applies Global Optimization, decomposing tasks and refining high-level plans through reflective analysis. This enables the system to dynamically adapt to evolving objectives while effectively managing the complexities of vast action spaces. Following this, WebPilot employs Local Optimization to execute each subtask using a customized MCTS approach, addressing uncertainties and incomplete information by iteratively refining decisions based on new observations.

Specifically, the Global Optimization phase is driven by \textit{Planner}, \textit{Controller}, and \textit{Extractor}. It begins with Hierarchical Task Decomposition (HTD), where \textit{Planner} breaks down complex tasks into manageable subtasks, narrowing the focus and effectively addressing the vast action spaces that challenge classical MCTS. Reflective Task Adjustment (RTA) then refines the high-level plan based on new observations, allowing WebPilot to adapt dynamically. \textit{Controller} monitors subtask progression, assessing subtask completeness and generating reflections if the subtask requires re-execution, ensuring accurate and adaptive task completion. Throughout this process, \textit{Extractor} gathers essential information to support task execution. This coordinated approach ensures WebPilot remains adaptable and efficient in dynamic environments.

For each subtask, WebPilot employs Local Optimization strategies driven by the \textit{Explorer}, \textit{Verifier}, \textit{Appraiser}, and \textit{Controller} to enhance execution in dynamic environments. Goal-Oriented Selection (GOS) harnesses the intuitions of LLMs to efficiently steer WebPilot toward the most promising states for subtask completion. Reflection-Enhanced Node Expansion (RENE) uses real-time feedback to continuously refine tactical-level strategies as conditions evolve. Dynamic Evaluation and Simulation (DES) continuously assesses actions and anticipates potential outcomes by integrating real-time feedback with one-step forward simulations. Maximal Value Backpropagation (MVB) prioritizes promising paths by focusing on strategies with the highest potential, updating values based on maximum future rewards. By integrating Local Optimization with Global optimization strategies, WebPilot ensures adaptable task execution, harnessing the specialized abilities and responsibilities of multiple agents to outperform existing web agents in dynamic environments.

Experiments on MiniWoB++ \cite{liu2018reinforcement} and WebArena \cite{zhou2023webarena} are chosen to assess the performance of WebPilot in environments with varying complexity. The results highlight the superiority of WebPilot, particularly in the complex, realistic web environment. In WebArena, WebPilot achieves an impressive 37.2\% success rate, surpassing the current SOTA method, SteP \cite{sodhi2024step}, which relies on rigid, expert-designed policies tailored to specific states and actions. Notably, WebPilot demonstrates a remarkable 93\% relative increase in success rate over concurrent tree-based methods \cite{koh2024treeCMU}. Even when using GPT-3.5, WebPilot remains highly competitive with GPT-4-based SOTA methods, achieving a 29.1\% success rate. These results underscore the exceptional ability of WebPilot to handle the uncertainty and complexity inherent in real-world web environments.

The primary contributions of this work are as follows:

1. We introduce WebPilot, an autonomous multi-agent system designed for complex web environments, combining global and local MCTS-inspired optimization strategies to enable human-like flexibility in exploration, adaptation, and decision-making at both the subtask and action levels.
	
2. We develop a Hierarchical Reflection Mechanism, incorporating Strategic Reflection in Global Optimization and Tactical Reflection in Local Optimization, which significantly enhances adaptive learning and decision-making in evolving environments.

3. We introduce a novel Granular Bifaceted Self-Reward Mechanism that guides MCTS by integrating action effectiveness with goal-oriented potential, allowing for more precise assessments in dynamic and ambiguous environments.

4. WebPilot achieves SOTA performance, particularly in challenging benchmarks like WebArena, demonstrating substantial advancements in general autonomous agent capabilities for complex real-world tasks.

\section{Related Work}
 In this section, we provide a comparative analysis of LLM-based web agents and MCTS-based agents, with a detailed comparison presented in Tab. \ref{table:benchmark-comparison}. For a more in-depth analysis of these agents, please refer to the Appendix.
 
\subsection{LLM-Based Autonomous Web Agents}
Recent advancements in LLMs have paved the way for the development of web agents that leverage the reasoning abilities of LLMs to interact with web environments. One line of LLM-based web agent \cite{kim2024languageRCI, sun2024adaplanner, prasad2023adapt, fu2024autoguide, ma2023laser, zheng2023synapse, tao2023webwise} predominantly relies on environment-specific state-action pairs embedded within demonstrations to respond to specific observations. For instance, SteP \cite{sodhi2024step}, currently the SOTA on WebArena\cite{zhou2023webarena}, utilizes rigid, expert-designed policies tailored to particular states and actions. However, these agents often struggle with exploring and adapting to realistic, unseen web tasks. Another line of web agents \cite{li2023azeroshot, zhou2023language, pan2024autonomous, koh2024treeCMU} adopts a strategy of freely exploring and discovering unknown environments. Despite efforts like Auto Eval \& Refine \cite{pan2024autonomous}, which incorporates an evaluator into Reflexion \cite{shinn2024reflexion}, and LM-Tree Search \cite{koh2024treeCMU}, which employs a search-based method in realistic environments, these agents still encounter challenges with complex tasks, leaving room for significant advancements. WebPilot, in contrast, is a multi-agent system employing a dual optimization strategy, excelling in exploring unseen tasks and dynamically adjusting strategies and actions based on new observations. This capability enables WebPilot to demonstrate superior adaptability in more complex environments.

\subsection{LLM-MCTS Applications}
MCTS, originally developed for the game of Go \cite{coulom2006efficient, browne2012survey}, is renowned for its effectiveness in handling exploration problems. Enhanced by the Upper Confidence bounds applied to Trees (UCT) method \cite{kocsis2006banditUCT}, MCTS has found extensive use in fields such as robotics\cite{zhao2024large}, strategy games \cite{jang2021monteFictionGame}, and autonomous vehicles \cite{lenz2016tactical}. Recently, researchers have integrated LLMs with MCTS to tackle various NLP tasks, including QA \cite{Hong2023FaithfulQA, xie2024monte, chi2024thoughtsculpt}, prompt refinement techniques \cite{Wang2023PromptAgentSP}, and complex mathematical reasoning problems \cite{Tian2024TowardSO, Zhang2024AccessingGL}. Building on this integration, LLM-based agents have also incorporated MCTS to enhance their exploratory and decision-making capabilities. For instance, LATS \cite{zhou2023language} applies MCTS to simple web tasks. However, traditional MCTS-based methods often encounter difficulties in scenarios with vast action spaces, unpredictable state transitions, and incomplete information in web tasks. WebPilot addresses these challenges by utilizing a tailored MCTS designed specifically for complex environments, effectively navigating and optimizing decision-making processes even in highly uncertain situations.

\section{Methodology}
In §\ref{Promblem Formulation}, we formally describe the web exploration task, highlighting the challenges posed by the uncertain nature and dynamic of these environments. To overcome these challenges, WebPilot employs a dual optimization strategy, i.e., \textbf{Global Optimization} and \textbf{Local Optimization}. During the Global Optimization phase, as detailed in §\ref{global}, WebPilot generates high-level plans and continuously refines these plans through reflective analysis. This is followed by the Local Optimization phase, described in §\ref{local}, where WebPilot engages in low-level strategic exploration. The entire process is outlined in Algo. \ref{algorithm}, with additional details provided in the Appendix.

\subsection{Promblem Formulation}
\label{Promblem Formulation}
Our objective is to enable the LLM-based web agent to effectively solve a task $\mathcal{T}$ in a web environment $\mathcal{E}$ by emulating human web navigation strategies. Web environments are inherently partially observable, which limits the information available to agents and complicates problem-solving. This partial observability occurs because web content can change dynamically, meaning the agent cannot fully anticipate or know the state of certain elements—such as updated content or availability—until it interacts with them. Consequently, agents must often make decisions under conditions of uncertainty and incomplete information. Following WebArena \cite{zhou2023webarena}, we use an accessibility tree, referred to as actree, to represent the observation, which captures the structure and interactive elements of the web page. However, due to the lack of specific web domain knowledge, LLMs often struggle to recognize or utilize the functionalities of various web elements. As a result, the agent must actively explore the environment to gather critical information about both the task and the functionality of the web elements, making informed decisions despite these uncertainties and incomplete information.

Specifically, this process can be modeled as a Partially Observable Markov Decision Process (POMDP). The environment $\mathcal{E}$ is defined by a state space $\mathcal{S}$, an action space $\mathcal{A}$, and an observation space $\mathcal{O}$. The transition function $\mathcal{F}: \mathcal{S} \times \mathcal{A} \rightarrow \mathcal{S}$ dictates how states evolve based on actions taken, typically in a deterministic manner governed by the environment. Task execution requires the agent to make decisions based on partial observations $o_t$ at each time step $t$. Each action $a_t$ results in a new state $s_{t+1}$ and an updated observation $o_{t+1}$. The evaluation function \(eval(\boldsymbol{a}, \boldsymbol{s})\), defined by the environment, assesses the success of task execution. Here, \(\boldsymbol{a} = \{a_1, \ldots, a_n\}\) represents the sequence of executed actions, and \(\boldsymbol{s} = \{s_1, \ldots, s_n\}\) denotes the corresponding sequence of intermediate states. This function evaluates whether the state transitions satisfy the criteria established by the task $\mathcal{T}$.

\begin{figure*}
    \centering
    \includegraphics[width=1\linewidth]{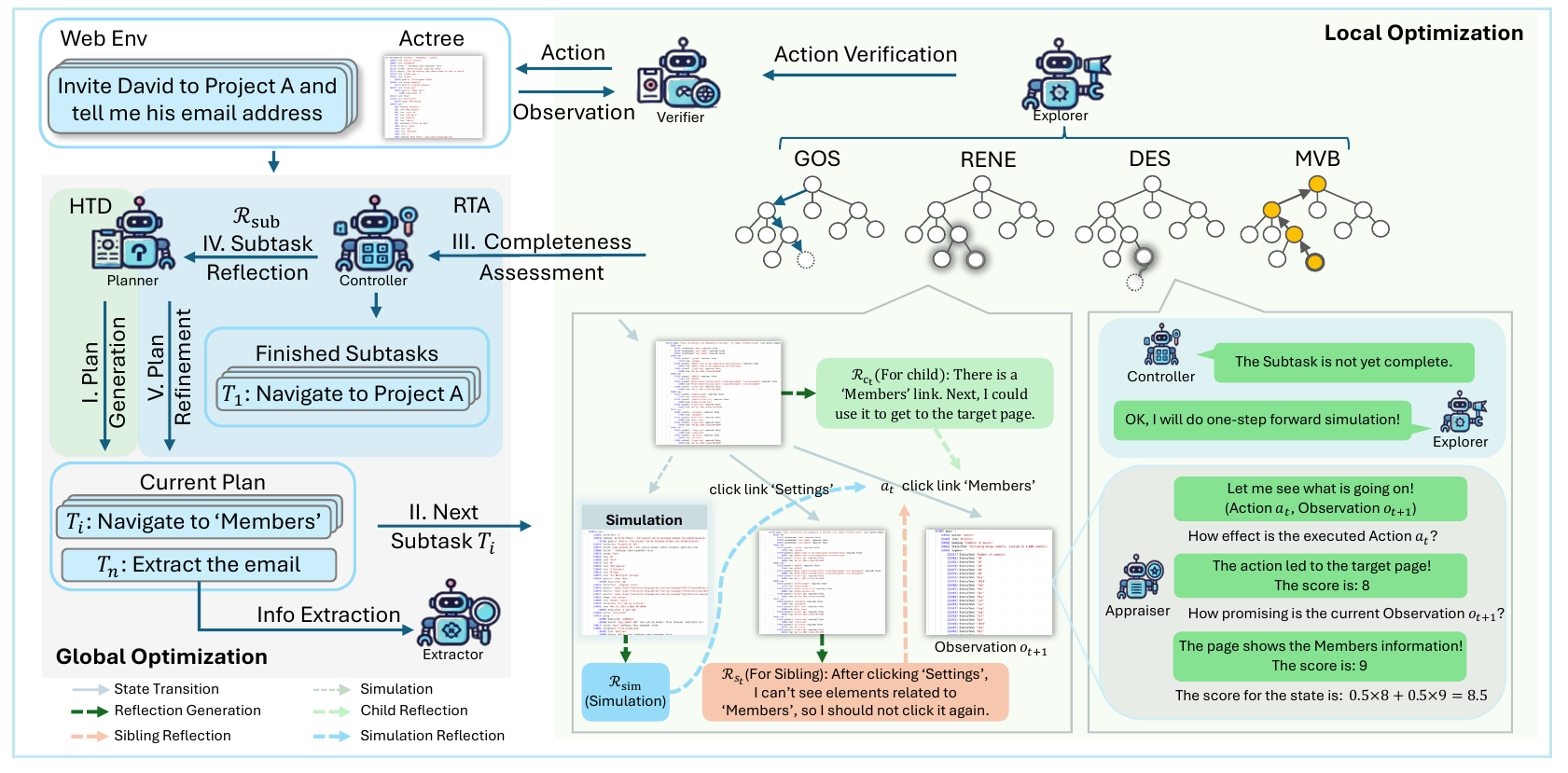}
    \caption{An overview of WebPilot. GOS: Goal-Oriented Selection; RENE: Reflection-Enhanced Node Expansion; DES: Dynamic Evaluation and Simulation; MVB: Maximal Value Backpropagation; HTD: Hierarchical Task Decomposition; RTA: Reflective Task Adjustment.}
    \label{fig:enter-label}
    \vspace{-10pt}
\end{figure*}

 \subsection{Global Optimization: Adaptive Strategy Refinement through Reflective Adjustment}
\label{global}
The Global Optimization phase emulates human cognition by leveraging prior knowledge to generate an initial plan for unfamiliar tasks. However, due to the lack of specific web domain knowledge in LLMs and the dynamic, uncertain nature of web environments, this initial plan misses critical details and struggles to remain effective as the environment evolves. To address this, WebPilot continuously refines the initial plan through reflective analysis of new observations and previous subtask outcomes. Global Optimization involves two key components: Hierarchical Task Decomposition (HTD) and Reflective Task Adjustment (RTA), which are facilitated by the \textit{Planner}, \textit{Controller}, and \textit{Extractor}.

\subsubsection{Hierarchical Task Decomposition (HTD)} begins with \textit{Planner} breaking down complex tasks into smaller, manageable subtasks $\mathcal{T}_i$, thereby creating a flexible high-level plan that can adapt to the uncertain and ever-changing conditions of web environments. In generating this plan, \textit{Planner} utilizes only a few high-level demonstrations to ensure robust and adaptive task decomposition; an example of this is provided in the Appendix. This approach allows WebPilot to dynamically adjust its strategies for specific aspects of each task, making it more responsive to environmental changes. Unlike the concurrent search-based web agent \cite{koh2024treeCMU}, which struggles with complex tasks due to its expanding search space, HTD ensures that each subtask is more targeted and efficient. This decomposition enables WebPilot to refine each subtask in real time, adjusting dynamically to evolving conditions without requiring a complete overhaul of the entire task execution. By concentrating on these manageable subtasks, WebPilot employs MCTS-enhanced decision strategies, specifically through the Local Optimization phase as described in §\ref{local}, to minimize unnecessary search paths and optimize decision-making within a focused scope, effectively mitigating the challenges posed by vast action spaces that often hinder classical MCTS. The effectiveness of \textit{Planner} is demonstrated through ablation studies, detailed in §\ref{ablation}.

\subsubsection{Reflective Task Adjustment (RTA)} Upon completing each subtask in the Local Optimization phase, as will be discussed in §\ref{local}, WebPilot reassesses and refines its high-level plan to ensure alignment with the overall task $\mathcal{T}$. Guided by \textit{Controller} and \textit{Planner}, this process critically evaluates the execution of each subtask against expected outcomes, allowing WebPilot to recalibrate its strategy based on new observations. \textit{Controller} plays a crucial role in this process by assessing whether the current observation $o_t$ and the executed action sequences \(\boldsymbol{a}\) align with the subtask $\mathcal{T}_i$. It then generates a subtask completeness $\text{Comp}_t$. If the completeness assessment indicates that the subtask is not complete, \textit{Controller} initiates a re-execution of the subtask. Before this re-execution, the $\text{Comp}_t$, along with the associated observation and executed actions, is used to generate a Strategic Reflection, i.e., subtask reflection $\mathcal{R}_{\text{sub}}$. This reflection guides the repeated execution of the subtask, leveraging the experience from the previous attempt to avoid repeating the same errors. Meanwhile, \textit{Extractor} continuously gathers critical information to support the successful completion of the task. An example of \textit{Controller} handling task completeness and subtask reflection is provided in the Appendix.

\begin{figure*}
    \centering
    \includegraphics[width=1\linewidth,height=7cm]{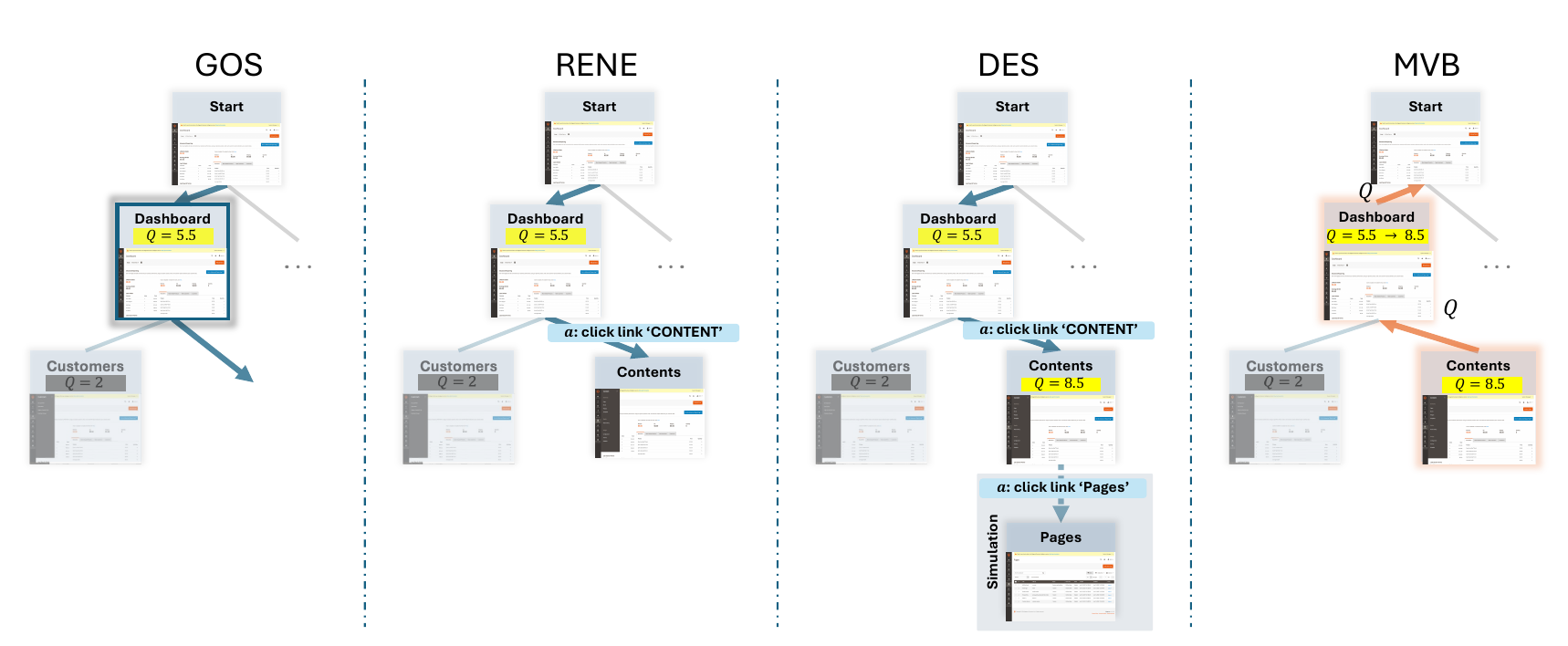}
    \caption{Overview of the Local Optimization stage in WebPilot, highlighting key components: Goal-Oriented Selection (GOS), Reflection-Enhanced Node Expansion (RENE), Dynamic Evaluation and Simulation (DES), and Maximal Value Backpropagation (MVB). The subtask in this example is “Navigate to the ‘Pages’ site.” For detailed results, please refer to the Appendix.}
    \vspace{-10pt}
    \label{fig:mcts_4_stages}
\end{figure*}

\subsection{Local Optimization: MCTS-Enhanced Decision Strategies}
\label{local}
The Local Optimization phase in WebPilot is inspired by the human-like adaptability required to navigate and solve complex web tasks, effectively captured by MCTS. For each subtask $\mathcal{T}_i$ and its subtask-specific goals $\text{Objective}_i$, which define the expected outcomes or milestones to be achieved within that subtask, \textit{Explorer}, \textit{Verifier}, and \textit{Appraiser} work together to complete the task. \textit{Explorer} identifies optimal actions, \textit{Verifier} ensures these actions are valid and non-redundant, and \textit{Appraiser} evaluates both the immediate effectiveness of an action and its potential to achieve the intended goal, offering continuous feedback for a more nuanced and accurate assessment. Throughout this process, \textit{Controller} assesses whether the subtask is completed and determines if further actions are needed, ensuring alignment with the overall task.

The Local Optimization phase of WebPilot, akin to classical MCTS, follows four key stages, as shown in Fig. \ref{fig:mcts_4_stages}: 1) Goal-Oriented Selection leverages the initial intuitions of LLMs to steer WebPilot toward the most promising paths for subtask completion, emulating how humans use prior knowledge to navigate tasks. 2) Reflection-Enhanced Node Expansion integrates feedback after each node expansion, enabling WebPilot to reassess and refine its strategy dynamically, much like human reflection informs decision-making. 3) Dynamic Evaluation and Simulation allows WebPilot to assess current states by analyzing executed actions and simulating potential outcomes, mirroring human foresight in evaluating consequences. 4) Maximal Value Backpropagation prioritizes long-term potential by continuously updating value estimates based on the maximum future rewards. Through this comprehensive Local Optimization phase, WebPilot effectively balances exploration and exploitation, leading to more efficient and informed decision-making in complex tasks.

\subsubsection{Goal-Oriented Selection (GOS)} directs \textit{Explorer} toward high-value nodes by leveraging the initial intuitions derived from the LLM. Although these intuitions are not specifically tailored to the environment, they provide valuable insights that effectively narrow the action space. These insights arise from the extensive pre-existing knowledge base of LLM, enabling it to make informed estimates about which actions are likely to be more promising, even without explicit training in the specific web domain. GOS employs a modified version of the PUCT selection method, inspired by AlphaGo \cite{Silver2017MasteringTG}:
\begin{equation}
    U(s,a) = w_{puct} \frac{\sqrt{\sum_bN(s,b)}}{1+N(s,a)},    
\label{puct}
\end{equation}
where \(w_{puct}\) represents the exploration bias factor balancing the exploration and exploitation, and \(N(s,a)\) represents the total count of conducting the action \(a\) in state \(s\). 

Unlike classical MCTS, as deployed in RAP \cite{hao-etal-2023-reasoning}, which prioritizes unexplored nodes due to the infinite potential assigned by traditional UCT \cite{kocsis2006banditUCT}, GOS refines this approach to better manage vast action spaces. The initial insights of LLM often assign high value to the first visited node based on its broad knowledge base. In contrast to UCT, which would mandate exploring unvisited nodes even when the current node is nearly optimal, GOS modifies the selection formula by adding a \text{+1} term in the denominator. This adjustment enables GOS to direct the \textit{Explorer} toward the most promising paths, minimizing unnecessary exploration and efficiently navigating complex environments, similar to how humans use prior knowledge in decision-making.

\subsubsection{Reflection-Enhanced Node Expansion (RENE)} navigates the vast action spaces of the web by integrating reflective feedback, i.e., Strategic Reflection and Tactical Reflection, at each step,  enabling WebPilot to continuously refine its strategy, enhancing decision-making through focused exploration and exploitation.

Specifically, to explore the state space efficiently, WebPilot departs from traditional MCTS by generating and expanding only one action per step. \textit{Explorer} generates actions $a_t$ and corresponding intents $\mathcal{I}_t$ in real time, continuously adjusting to evolving conditions with guidance from reflective feedback. \textit{Verifier} ensures that the action $a_t$ is both valid and unique among sibling nodes. Formally,
\begin{equation}
    a_t, \mathcal{I}_t = \textit{Explorer} (o_t, \mathcal{T}_i, H_t, \mathcal{R}_t, \mathcal{C}_{t-1}),
\end{equation}
where \(\mathcal{T}_i\) the subtask, and \(H_t=\{a_{1},\ldots, a_{t-1}\}\) the action history. If available, the Tactical Reflections—comprising simulation reflection $\mathcal{R}_{\text{sim}_t}$, parent reflection $\mathcal{R}_{p_t}$, and sibling reflection $\mathcal{R}_{s_t}$—along with the Strategic Reflection, which is the subtask reflection $\mathcal{R}_{\text{sub}}$, are incorporated into $\mathcal{R}_t = \{\mathcal{R}_{\text{sim}_t}, \mathcal{R}_{p_t}, \mathcal{R}_{s_t}, \mathcal{R}_{\text{sub}}\}$ and used together with the continuation reason $\mathcal{C}_{t-1}$ to leverage prior exploration. This combined use of available reflections helps narrow the action space and optimize the MCTS process.

Upon executing $a_t$, the environment returns the resulting state and observation $(s_{t+1}, o_{t+1})$.  \textit{Explorer} then compares the current observation $o_{t+1}$ with the previous one $o_t$ to determine whether the action intent $\mathcal{I}_t$ has been achieved: 
\begin{equation}
    \text{Effect}(a_t) = \textit{Explorer}(o_{t+1}, o_{t}, \mathcal{I}_t)
\end{equation}
Following this evaluation, Child Reflection $\mathcal{R}_{c_t}$ and Sibling Reflection $\mathcal{R}_{s_t}$ are generated:
\begin{equation}
    \mathcal{R}_{c_t}, \mathcal{R}_{s_t} = \textit{Explorer}(\text{Effect}(a_t), \mathcal{T}_i, \text{Objective}_i, o_t, H_t),
\end{equation}
where \(\text{Effect}(a_t)\) captures the impact of \(a_t\) on the current state, and \(\text{Objective}_i\) denotes the subtask-specific goals, which define the expected outcomes or milestones to be achieved within that subtask $\mathcal{T}_i$.  Detailed descriptions of how $\text{Effect}(a_t)$ is determined are provided in the Appendix.

\begin{figure}
    \centering
    \includegraphics[width=1\linewidth]{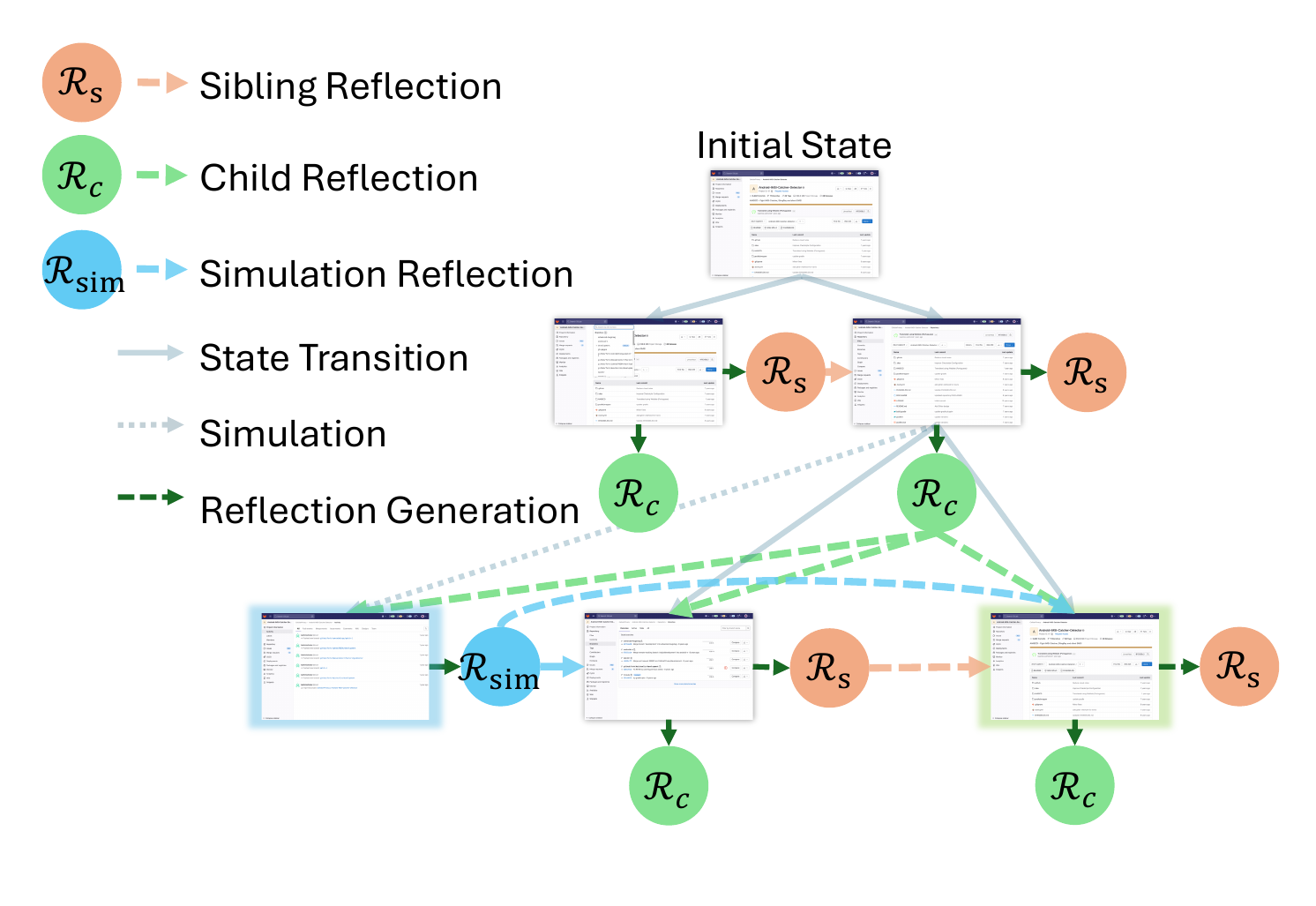}
    \caption{Tactical Reflection Flow. Diagram illustrating how Child, Sibling, and Simulation Reflections guide decision-making and exploration in WebPilot.}
    \label{reflection_flow}
    \vspace{-10pt}
\end{figure}

Child Reflection guides the generation of the next action, with $\mathcal{R}_{c_t}$ becoming the parent reflection $\mathcal{R}_{p_{t+1}}$ for the child node, ensuring continuity in the decision-making process. This continuity is crucial for complex tasks, where maintaining a coherent reasoning path is essential. Disruptions in this flow can significantly impair performance, as demonstrated in §\ref{ablation}. Sibling Reflection $\mathcal{R}_{s_t}$ enhances exploration by leveraging insights from previous sibling node explorations, enabling the agent to focus on promising areas and uncover new possibilities when encountering similar scenarios. Together, Child Reflection and Sibling Reflection help WebPilot narrow the vast action space to a more manageable subset when generating actions. By integrating feedback from reflections on already executed actions, WebPilot effectively analyzes transitions and assesses decisions made under incomplete information, thereby enhancing its overall efficiency and performance in dynamic environments. The roles and impacts of these reflection mechanisms are further detailed in §\ref{ablation}, with illustrative examples of these reflections provided in the Appendix. For a comprehensive view of the Tactical Reflection, see Fig. \ref{reflection_flow}.

\subsubsection{Dynamic Evaluation and Simulation (DES)} dynamically assesses how generated actions align with evolving task states by leveraging real-time feedback rather than relying on predefined reward structures typical of classical MCTS. This adaptive evaluation ensures that each action remains responsive to the changing environment, guiding the agent toward the evolving goal.

The reward function is crucial in MCTS, but prior methods often rely on direct feedback from the environment \cite{zhou2023language}, which is impractical for realistic web tasks, or use binary success/failure outcomes or simplistic intermediate states \cite{koh2024treeCMU}. Such approaches frequently misjudge the ambiguous and evolving nature of web environments, leading to inaccurate evaluations. Moreover, intermediate steps on the web are challenging to categorize as simply correct or incorrect because their effectiveness in contributing to the final task outcome may not be immediately apparent. Inspired by the A* algorithm \cite{4082128Astar}, \textit{Appraiser} evaluates both the effectiveness of the executed action $a_t$ and the potential of the resulting observation $o_{t+1}$ to achieve the intended goal, providing a more nuanced and dynamic assessment. This approach is refined using a granular 0-10 scoring system, which allows for a more precise evaluation of action impact, capturing the subtleties of evolving and uncertain web environments.

This Granular Bifaceted Self-Reward Mechanism evaluates both the effectiveness of the action taken, $S_{\text{eff}}(a_t)$, and the potential of the resulting observation, $S_{\text{fut}}(o_{t+1})$, using a precise 0-10 scoring system. This approach allows for a more nuanced assessment, capturing subtle differences in action quality and future potential, which is crucial for determining whether to proceed with the current strategy. Illustrative examples of this novel mechanism are provided in the Appendix.
\begin{equation}
S_{\text{eff}}(a_t), S_{\text{fut}}(o_{t+1}) = \textit{Appraiser}(\text{Effect}(a_t), o_{t+1}, \mathcal{T}_i),
\end{equation}
where \textit{Apprasier} assesses how well the state aligns with the subtask \(\mathcal{T}_i\), and $\text{Effect}(a_t)$ represents the change between the former and current state.

The overall reward $S_{\text{total}} $ aggregates these scores to represent the value of the current state:
\begin{equation}
S_{\text{total}}(a_t, o_{t+1}) = w_{\text{eff}} \cdot S_{\text{eff}}(a_t) + w_{\text{fut}} \cdot S_{\text{fut}}(o_{t+1}),
\end{equation}
where \(w_{\text{eff}}\) and \(w_{\text{fut}}\) balance action effectiveness and future potential.

After evaluating the current state, \textit{Controller} determines whether the subtask \(\mathcal{T}_i\) is complete,
\begin{equation}
    \mathcal{C}_t = \textit{Controller}(\mathcal{T}_i, \{a_1, a_2, \ldots, a_t\}, o_{t+1}),
\end{equation}
where \(\mathcal{C}_t\) is a continuation reason based on the executed actions and current observation. If the subtask is confirmed complete, the search terminates. Otherwise, DES conducts a one-step forward simulation guided by RENE, generating a Simulation Reflection $\mathcal{R}_\text{sim}$, which acts as a shallow trial to provide insights for the next exploration. This process enables the agent to better understand unpredictable transitions by simulating potential outcomes, which helps clarify ambiguities and uncertainties. The quantitative values assigned through these simulations guide the agent in making more informed decisions, effectively navigating the most promising paths while managing unpredictability and incomplete information.

\subsubsection{Maximal Value Backpropagation (MVB)} enhances the traditional MCTS backpropagation step by prioritizing the most promising paths. Unlike classical MCTS, which typically averages the values of child nodes and may lead to the exploration of less optimal paths, MVB uses the maximum value from all child nodes, $\max Q(s_{t+1})$, where $Q(s)$ represents the potential value for completing the subtask in state $s$. For the first visited state, the Q-value is initialized to the score $\mathcal{S}_{\text{total}}$ as evaluated in DES. This approach accumulates values throughout the decision tree, consistently targeting strategies with the highest potential for long-term success. By focusing on these high-value paths, WebPilot ensures alignment with the ultimate goal rather than merely advancing to the next immediate step.

\begin{figure}[h] 
\centering
\scalebox{0.97}{ 
    \parbox{\linewidth}{ 
    \begin{algorithm}[H] 
    \caption{WebPilot}
    \label{algorithm}
    \begin{algorithmic}[1]
    \REQUIRE task $\mathcal{T}$; max node count $n_{\text{max}}$
    \STATE \textbf{HTD}: \textit{Planner} generates a plan $\mathcal{P} = \{\mathcal{T}_1, \mathcal{T}_2, \ldots, \mathcal{T}_n\}$
    \WHILE{$\mathcal{P}$ is not empty}
        \STATE Execute the first subtask $\mathcal{T}_1$ from $\mathcal{P}$
        \IF{$\mathcal{T}_1$ involves information extraction}
            \STATE \textit{Extractor} processes the current observation $o_t$
        \ELSE
            \WHILE{$n_{\text{max}}$ is not reached}      
                \STATE \textbf{GOS}: Select the target node via Eq. \hyperref[puct]{\ref{puct}}
                \STATE \textbf{RENE}: \textit{Explorer} expands the search tree to the next state $(s_{t+1}, o_{t+1})$, observes the environment, and generates reflections $\mathcal{R}_{c_t}, \mathcal{R}_{s_t}$
                \IF{\textit{Controller} decides to stop the subtask $\mathcal{T}_1$}
                    \STATE \textbf{break}
                \ELSE
                    \STATE \textit{Controller} generates continuation reason $\mathcal{C}_t$
                \ENDIF
                \STATE \textbf{DES}: \textit{Appraiser} evaluates the current state $S_{\text{total}}(a_t,o_{t+1})$, and \textit{Explorer} conducts one step forward simulation to generate simulation reflection $\mathcal{R}_{\text{sim}}$
                \STATE \textbf{MVB}: Backpropagate the reward of the expanded node
            \ENDWHILE
        \ENDIF
        \STATE \textbf{RTA}: \textit{Controller} assesses the completeness of the subtask $\mathcal{T}_1$, generates subtask reflection $\mathcal{R}_{\text{sub}}$ if needed and updates the current plan $\mathcal{P}$
    \ENDWHILE
    \end{algorithmic}
    \end{algorithm}
        \vspace{-10pt}
    }
    \vspace{-10pt}
}
\end{figure}

\section{Experiment}
\subsection{Experimental Setup}

\subsubsection{Datasets and Metrics}
To demonstrate the broad applicability of WebPilot, we evaluate our method on two benchmarks: WebArena \cite{zhou2023webarena} and MiniWoB++ \cite{liu2018reinforcement}. WebArena, comprising 812 human-annotated web tasks, is designed to assess the ability of agents to perform actions on complex, realistic websites. These tasks are diverse, long-horizon, and closely mirror the types of activities humans routinely engage in online. Notably, WebPilot operates as a text-only agent, leveraging the accessibility tree, i.e., actree, of the webpage without relying on visual observations—a limitation we aim to address in future work. WebArena is chosen as the primary benchmark due to its close simulation of real-world web environments. Additionally, we validate WebPilot on MiniWoB++, an environment offering a range of simpler yet varied web tasks, from basic interactions like button-clicking to more complex activities such as form-filling, which require reasoning capabilities. For evaluation, we use the Success Rate (SR) metric as defined in \cite{zhou2023webarena} for WebArena, and follow \cite{li2023azeroshot} for MiniWoB++, focusing on 43 tasks that can be completed solely through text representation.

\subsubsection{Baseline Models}
We compare WebPilot against several baseline models, including the current SOTA SteP \cite{sodhi2024step} and the concurrent search-based model LM-Tree Search (LM-TS) \cite{koh2024treeCMU}. We utilize GPT-3.5 and GPT-4o \footnote{GPT-3.5-turbo-0125, GPT-4o-2024-05-13}, each configured with a max\_tokens limit of 4096 tokens, and a temperature setting of 0.3, while all other parameters are kept at their default values. 

\subsubsection{Implementation Details}
We run WebPilot with parameters optimized for both efficiency and performance. Specifically, we limit the max node count per subtask to 10 and set the exploration bias at 5 to balance exploration and exploitation. Refer to the Appendix for details. 

\subsection{Results on WebArena}
As shown in Tab. \ref{webarena_performance}, WebPilot consistently outperforms existing methods. In WebArena, WebPilot with GPT-4o demonstrates a remarkable 93\% relative increase in SR compared to concurrent tree search-based methods LM-TS, achieving a 37.2\% SR. This performance surpasses the current SOTA method, SteP \cite{sodhi2024step}, which relies on rigid, expert-designed policies tailored to specific states and actions. This substantial improvement highlights the effectiveness of the adaptable and dynamic approach of WebPilot in navigating complex web environments.

\paragraph{Greater Flexibility and Adaptability in WebPilot}
Compared with SteP, WebPilot demonstrates a significant 7.7\%  improvement in the Gitlab domain. This advantage arises from the strategic use of high-level demonstrations, which equip the agent with general web domain knowledge rather than confining it to rigid, expert-designed policies specific to certain states and actions, as SteP does. The Gitlab domain, characterized by its diverse and complex tasks, as well as dynamic, multi-step scenarios, highlights the ability of WebPilot to generalize knowledge and adapt strategies in real time. This broader understanding allows WebPilot to more effectively infer and address unseen tasks. Additionally, the exploratory approach of WebPilot, which mirrors human adaptability, facilitates dynamic navigation and problem-solving in unfamiliar scenarios, further enhancing its overall performance in uncertain web environments.

\paragraph{Superior Task Performance in WebPilot Through Strategic Decomposition and Reflective Feedback}
WebPilot significantly outperforms the search-based method, LM-TS, particularly in the Reddit and GitLab domains, due to two key factors. First, the lack of strategic decomposition by the \textit{Planner} in LM-TS makes navigating the vast state space considerably more challenging, as the agent lacks clear guidance, leading to less efficient exploration and poorer performance. The importance of this decomposition is highlighted by the superior results achieved by WebPilot, as demonstrated by the ablation studies in §\ref{ablation}. Second, WebPilot employs hierarchical reflections after each node expansion, enabling continuous reassessment and refinement of its strategy. Further details can be found in §\ref{ablation}.

\paragraph{Enhanced Reasoning and Planning Capabilities Remain Crucial for Improvement} 
Even with the less powerful GPT-3.5, WebPilot demonstrates a significant improvement over the WebArena baseline, highlighting its effectiveness in leveraging MCTS-inspired strategies to navigate complex environments. However, transitioning from GPT-3.5 to GPT-4o yields substantial gains, particularly in the Shopping, Reddit, and GitLab, with SR increases of 11.8\%, 6.6\%, and 9.4\%, respectively. These improvements are largely driven by the enhanced reasoning and planning capabilities of GPT-4o, which are crucial for tasks requiring precise inference and information retrieval in Shopping, as well as for navigating the more complex and diverse environments of GitLab and Reddit. The advanced planning abilities and the capacity to generalize domain knowledge from limited demonstrations to unseen tasks enable WebPilot to generate more effective plans, better understand the environment, and execute accurate actions. This underscores the importance of addressing the core challenges that current LLMs face in reasoning and planning, suggesting that further enhancements can be achieved with more powerful LLMs. 

\begin{table}[ht]
\centering
\scalebox{0.8}{
\begin{tabular}{l|>{\centering\arraybackslash}p{0.5cm} >{\centering\arraybackslash}p{0.5cm} >{\centering\arraybackslash}p{0.5cm} >{\centering\arraybackslash}p{0.5cm} >{\centering\arraybackslash}p{0.5cm} >{\centering\arraybackslash}p{0.5cm}}
\toprule
\textbf{Method} & \textbf{CMS} & \textbf{Map} & \textbf{Shop.} & \textbf{Red.} & \textbf{Git.} & \textbf{Avg} \\  
\midrule
\multicolumn{7}{l}{\textit{GPT-3.5}} \\ 
\midrule
WebArena \cite{zhou2023webarena} & -  & -  & -  & -  & -  & 8.9  \\ 
WebPilot (Ours) & \textbf{22.0} & \textbf{30.3} & \textbf{25.1} & \textbf{58.5} & \textbf{30.0} & \textbf{29.1} \\
\midrule
\multicolumn{7}{l}{\textit{GPT-4o}} \\ 
\midrule
WebArena \cite{zhou2023webarena} & -  & -  & -  & -  & -  & 13.1  \\ 
SteP \cite{sodhi2024step} & 24.2 & 30.3 & \textbf{36.9} & 59.4 & 31.7 & 33.5 \\ 
LM-TS \cite{koh2024treeCMU} & 16.5 & 25.8 & 28.1 & 10.5 & 13.3 & 19.2 \\
WebPilot (Ours) & \textbf{24.7} & \textbf{33.9} & \textbf{36.9} & \textbf{65.1} & \textbf{39.4} & \textbf{37.2} \\
\bottomrule
\end{tabular}
\vspace{-10pt}
}
\caption{Performance comparison in WebArena. Values represent SR as percentages. Domains are abbreviated as: Shop. (Shopping), Red. (Reddit), Git. (Gitlab). SteP uses GPT-4. Best results are bolded. WebPilot achieves SOTA results, showing a 93\% relative increase over LM-TS.}
\label{webarena_performance}
\end{table}

\subsection{Results on MiniWoB++}
\begin{table}[ht]
\centering
\scalebox{0.9}{
\begin{tabular}{lc}
\toprule
\textbf{Method} & \textbf{SR}  \\ 
\midrule
RCI \cite{kim2024languageRCI} & 94.0\% \\
AdaPlanner \cite{sun2024adaplanner} &  92.9\% \\
A zero-shot \cite{li2023azeroshot} & 94.9\% \\ 
SteP \cite{sodhi2024step} & 96.0\% \\
WebPilot (Ours) & 95.6\% \\ 
\bottomrule
\end{tabular}
\vspace{-5pt}
}
\caption{Success rate (SR) on MiniWoB++. }
\vspace{-10pt}
\label{table:miniwob-comparison}
\end{table}

As shown in Tab. \ref{table:miniwob-comparison}, WebPilot achieves results competitive with the SOTA, SteP. The slight edge of SteP is due to its use of 10 action-level demonstrations, while WebPilot uses only 4 high-level demonstrations, leaving exploration to the agent. The simplicity of many MiniWoB++ tasks, which require minimal actions, also reduced the need for extensive exploration, limiting the advantage of WebPilot. Despite this, our method proved effective with far fewer demonstrations compared to other LLM-based agents. Detailed analysis of MiniWoB++ can be found in the Appendix.

\subsection{Ablation Studies} 
\label{ablation}
To evaluate the impact of each component in WebPilot, we categorized tasks into information-seeking (IS) and website interaction (WI), including site navigation and content configuration. IS tasks focus on extracting information, while WI tasks require executing complex action sequences. For the ablation experiments, we selected 50 IS tasks and 50 WI tasks where WebPilot successfully completed the objectives. This selection allowed us to focus on cases where the components are functioning optimally, providing a clearer understanding of how each component contributes to overall performance. The results, shown in Tab. \ref{ablation_results}, highlight the critical role each component plays in the effectiveness of WebPilot. Notably, WI tasks are more adversely affected than IS tasks, underscoring the importance of our design in equipping the agent to handle complex web exploration tasks. Relevant examples illustrating the following findings are provided in the Appendix.

\begin{table}[ht]
\centering
\scalebox{0.9}{
\begin{tabular}{r|cc}
\hline
& IS & WI\\ \hline
w/o Child Reflection    & 74\%                & 70\%                \\
w/o Sibling Reflection  & 72\%                & 60\%                \\
w/o \textit{Controller} & 86\%                & 60\%                \\
w/o \textit{Planner}    & 48\%                & 24\%                \\ \hline
WebPilot             & 100\%                & 100\%                \\ \hline
\end{tabular}
}
\vspace{-5pt}
\caption{Ablation studies on WebArena.}
\vspace{-15pt}
\label{ablation_results}
\end{table}

\paragraph{Importance of Child Reflection for Maintaining Coherent Thought Processes}
The Child Reflection mechanism is crucial for maintaining a coherent thought process during exploration, aligning with the principles of Chain-of-Thought reasoning \cite{wei2022chain}. This reflection ensures that when generating actions for child nodes, the model has access to the parent node behind previous actions and its intended next steps. This continuity enables the model to produce more accurate and contextually relevant actions, preserving the logical flow necessary for complex decision-making. Without Child Reflection, this coherence is disrupted, leading to a significant 30\% decline in performance, particularly in WI tasks, where maintaining a consistent thought process is essential for success.

\paragraph{Critical Role of Sibling Reflection in Effective and Diverse Exploration}
Sibling Reflection is key to optimizing exploration within MCTS and expanding into diverse, promising areas, particularly for complex tasks. By leveraging insights from previously explored sibling nodes, WebPilot reduces redundancy and focuses on high-potential paths, ensuring valuable solutions are not missed. This mechanism enhances exploration effectiveness, especially for complicated web interaction tasks, as evidenced by WI tasks being more affected, as shown in Tab. \ref{ablation_results}.

\paragraph{\textit{Controller} is Critical for Subtask Accuracy and High-Level Decision-Making}
\textit{Controller} is crucial for Global Optimization, enabling the reassessment and refinement of plans. After each subtask, \textit{Controller} evaluates its completeness and, in collaboration with \textit{Planner}, refines the overall plan. Without \textit{Controller}, the verification of subtask completion is significantly compromised, leading to a noticeable performance decline, particularly in WI tasks. The greater impact on these tasks, which involve more subtasks, is due to the higher likelihood of incomplete subtasks as their quantity grows. This underscores the vital role of \textit{Controller} in high-level strategic planning and refinement. This finding highlights the importance of trial-and-error, mimicking human problem-solving, where \textit{Controller} serves as a key link between overarching goals and detailed task execution.

\paragraph{\textit{Planner} is Essential for Maximizing WebPilot Performance in Complex Tasks}
The removal of \textit{Planner} significantly degrades performance, particularly in complex WI tasks, due to the task complexity and the high number of subtasks. This decline stems from the inherent difficulty of MCTS with complex tasks. Task decomposition enables MCTS to operate more effectively.

\subsection{Agent Behavioral Analysis}
\label{behavioral}
This section analyzes the behavior of WebPilot to show its effectiveness in mimicking human strategies, with examples provided below and additional details in the Appendix.

\paragraph{WebPilot Can Learn Website Functionality by Exploring Unknowns} When switching branches in GitLab, the model first tries a dropdown menu but fails due to too many options, reflecting its limited web knowledge. By exploring further, it adapts and successfully finds the correct branch using the Branch Option in the sidebar, demonstrating its ability to learn and navigate unfamiliar web environments.

\paragraph{WebPilot Can Continuously Adapt Strategies and Actions Based on New Observations} On certain pages, elements like "link Project A" and "statictext Project A" appear multiple times in the actree, referring to the same page. While interacting with the statictext yields no changes, clicking the link achieves the desired outcome. WebPilot observes the error with the statictext, adjusts, and shares this reflection with other nodes, enhancing future decisions and demonstrating its adaptive learning capacity.

\paragraph{WebPilot Can Resolve Ambiguities Through Iterative Refinement} When tasked with checking "Merge request assigned to me", the initial plan involves accessing the merge request and filtering results. However, during the first subtask, WebPilot discovers a more direct path to the goal. Upon successfully completing this subtask, \textit{Controller} can determine that the remaining subtasks have been implicitly accomplished, demonstrating the ability of WebPilot to iteratively refine its strategy to resolve ambiguities.

\subsection{Limitation}
Despite the substantial progress achieved with WebPilot, limitations remain that impact its performance and could be addressed in future work. First,  the effectiveness of WebPilot is limited by the capabilities of LLMs, particularly in accurately understanding and interacting with complex web environments via text-based actions. Second, the absence of visual information, which forces the LLM to infer context typically provided by visual cues, e.g., the number of stars associated with the item, places additional strain on its performance. These challenges highlight the need for advancements in LLMs and the integration of visual reasoning to better handle complex web tasks. Refer to the Appendix for details.

\section{Conclusion}
We introduce WebPilot, a multi-agent system with a dual optimization strategy to enhance the adaptability and effectiveness of autonomous agents in complex web environments. By combining global optimization with a tailored MCTS-based local optimization, WebPilot overcomes the limitations of rigid predefined policies, achieving SOTA results on WebArena. Our approach marks a significant step forward in general autonomous agent capabilities, paving the way for more advanced decision-making in complex environments.

\bibliography{aaai25}

\newpage
\clearpage
\appendix
\onecolumn
\tableofcontents

\newpage
\section{Extended Related Works}

Tab. \ref{table:benchmark-comparison} in the main paper presents a comparison of different agent types. This comparison evaluates these agents against WebPilot, focusing on key features such as Dynamic Interaction, Partially Observable Environments, Non-Fixed Policies, Scalability, Realistic Web Environments, and Comprehensive Self-Reward Functions.

RAP \cite{hao-etal-2023-reasoning} remains centered on NLP tasks, including Blockworlds, mathematical reasoning, and question answering. In these tasks, the system state is represented purely in text form, meaning RAP operates without requiring dynamic interaction and is not suited for partially observable environments. 
Its reward function is derived from the log probability of the output of LLMs, with self-evaluation focused solely on the current state. In contrast, our work focuses on designing an agent capable of interacting dynamically with the environment, particularly in addressing complex, dynamic real-world web environments. The key challenge in these scenarios is enabling the agent to understand the environment and comprehend the impact of its actions on that environment. Unlike RAP, which operates in a static text-based context, our approach emphasizes the importance of real-time interaction and adaptability, allowing the agent to navigate and respond effectively to the evolving state of the web environment.

LATS \cite{zhou2023language} targets tasks like question answering, WebShop, and programming. In the WebShop environment, it demonstrates dynamic interaction capabilities. However, LATS relies on direct feedback from the environment—using outcome rewards that indicate task success to compute Q-values, which guide MCTS in its decision-making process. In real-world applications, such outcome rewards are often unavailable or impractical to obtain. For example, in many web-based tasks, the immediate success of an action may not be evident, or the outcome may depend on complex, long-term factors such as user behavior or the cumulative effects of multiple actions. To overcome this challenge, our approach avoids relying on outcome rewards. Instead, we employ adaptive strategies that allow the agent to learn and adjust its actions based on intermediate feedback during the task execution. This involves using real-time observations and ongoing assessments to refine decisions dynamically, enabling the agent to navigate complex environments and make informed decisions without needing explicit outcome-based rewards.

LLM-MCTS \cite{zhao2024large} integrates LLMs with MCTS to plan actions for a home robot within a simulated environment. This method leverages predefined examples drawn from similar scenarios to guide decision-making for each action. These examples serve as a heuristic, limiting the ability of agents to dynamically adapt to new, unforeseen situations, as it primarily relies on historical data to inform decisions. In contrast, WebPilot, though informed by initial guidance through high-level demonstrations, enables the agent to actively explore and learn from the environment in real time, adjusting its plan to address unseen tasks, as detailed in §\ref{demo_generalize}. This allows the agent to dynamically adapt to unseen tasks and complex web environments, making it more versatile and capable of managing situations where prior examples are not directly applicable.

SteP \cite{sodhi2024step}, the SOTA method on WebArena, excels in executing tasks by dynamically composing a library of predefined policies, each equipped with dedicated instructions and state-specific action examples. These policies are stacked and invoked as needed to handle specific subproblems, making SteP effective for structured tasks. However, this reliance on a predefined library constrains its adaptability, particularly in novel or unforeseen scenarios where the predefined policies may not apply. In contrast, our approach, while informed by high-level demonstrations, does not depend on static state-specific action policy libraries. Instead, it empowers the agent to actively explore and learn from the environment in real time, dynamically adjusting its strategy to handle new and complex tasks. This flexibility allows our method to perform effectively in dynamic and uncertain web environments, where rigid, predefined policies may fall short.

LM-Tree Search \cite{koh2024treeCMU}, a concurrent method, performs tree search within web environments without task decomposition, leading to scalability challenges as the search tree can become excessively large in complex tasks. It also relies on sparse reward functions that focus solely on executed actions, often resulting in suboptimal performance in dynamic web environments where immediate success is not always clear. In contrast, our approach does not rely on sparse rewards. Instead, it uses a dynamic evaluation strategy that assesses how well actions align with the evolving task state by directly observing real-time feedback. This continuous evaluation, which considers both the immediate effectiveness of actions and their potential to achieve the final goal, allows our method to adapt more effectively to complex and uncertain environments, ensuring that the agent remains responsive to the nuanced and changing nature of web tasks.

\section{Experimental Setup}
To thoroughly assess the capabilities of WebPilot, we conduct experiments in WebArena and MiniWoB++. Note that the web task environment remains only partially observable. Certain task-relevant elements may not always be visible in the current view. For example, after several scroll actions when reviewing commit records, the current branch might no longer be visible, complicating the decision-making process of agents. Additionally, web content can change dynamically, meaning the agent cannot fully anticipate or know the state of certain elements—like updated content or availability—until it interacts with them.

\subsection{Challenges in WebArena}
\label{Challenges in WebArena}

WebArena, derived from real-world websites, serves as a primary benchmark due to its complexity and realism. This environment presents several challenges that test the capabilities of WebPilot, each corresponding to a different aspect of human cognitive flexibility and adaptability.

\subsubsection{Similarity of Elements and Unknown Web Functionality}

\begin{wrapfigure}{r}{0.26\linewidth}
    \centering
   \includegraphics[width=\linewidth,height=3.8cm]{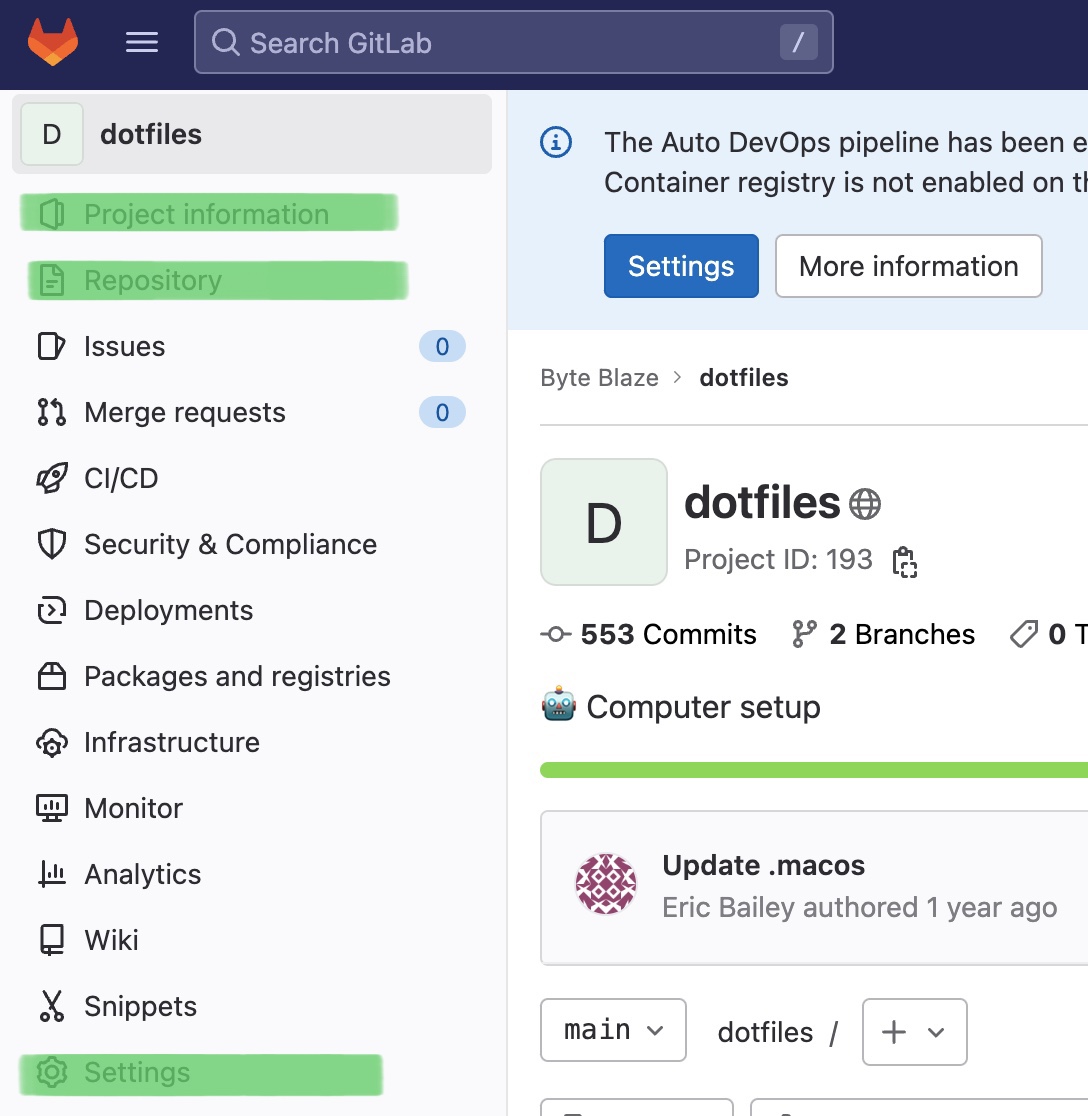}
    \caption{Multiple elements have similar meanings.}
    \vspace{-10pt}
    \label{webarena_hard}
\end{wrapfigure}
Realistic web environments, such as those modeled in WebArena, often present challenges due to the similarity of elements and the limited understanding the agent has of specific web functionalities. As illustrated in Fig. \ref{webarena_hard}, many elements in the sidebar, such as "Repository" and "Project Information," appear semantically related to links like "Members," which can lead the agent to rely on semantic guessing to determine the correct path. Additionally, the functionality of web elements may not be immediately clear, particularly when the agent encounters unfamiliar layouts or options. In response to this challenge, WebPilot first clicks the "Repository" link to see if the link to the "Members" page would appear, but upon finding nothing, it generates a reflection suggesting that it should interact with other elements and trace back to the previous state. The agent then clicks the "Project Information" link to perform the same check, and fortunately, it finds the entrance to the "Members" page.

\subsubsection{Complexity of Web Elements in Accessibility Tree, i.e., actree}
\begin{figure}
    \centering
    \includegraphics[width=0.95\linewidth]{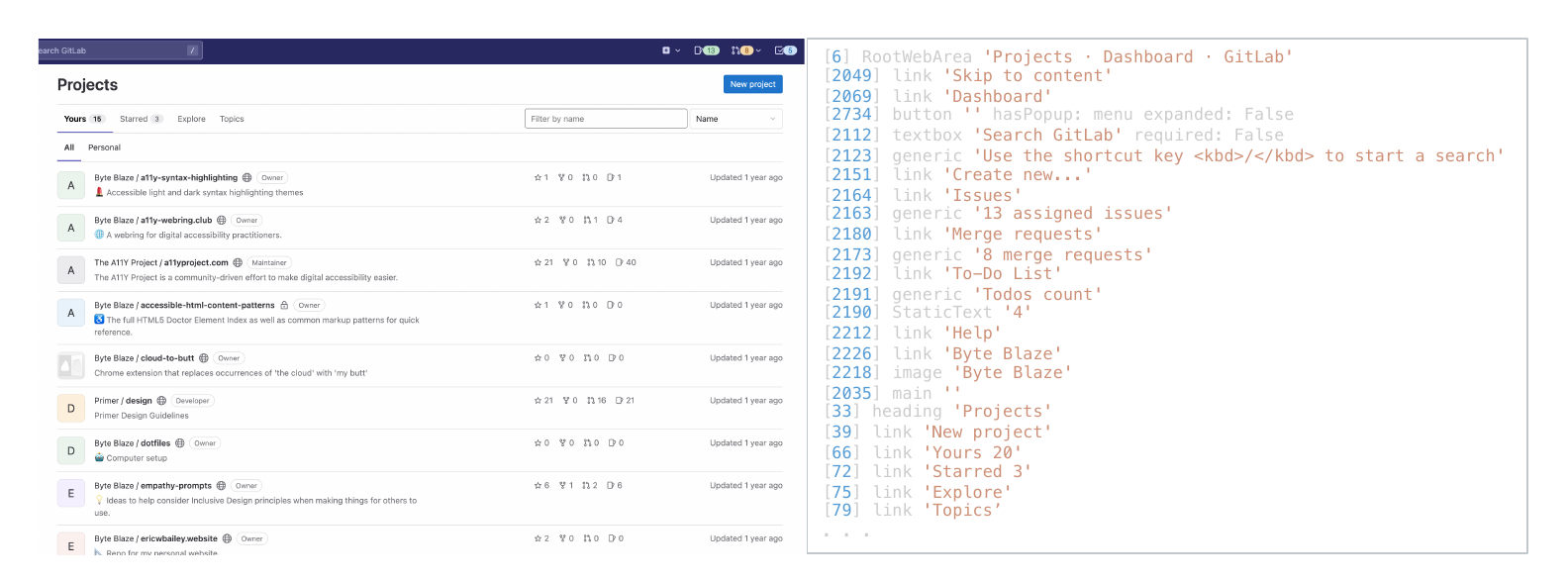}
    \caption{Comparison between the screenshot and the actree of the start page in Gitlab}
    \label{actree_example}
\end{figure}

The attempt to faithfully reconstruct the environment from HTML in WebArena results in the accessibility tree, i.e., actree, containing many elements with multiple representations, some of which are non-interactive. For example, elements like "link Project A" and "statictext Project A" may appear multiple times in the actree, both referring to the same page. While interacting with the statictext does not lead to any changes, clicking the link achieves the desired outcome. Even when visual information is available, the ambiguity between similar elements remains a challenge, as the visual cues do not always clearly distinguish functional differences between these elements. When encountering such situations, WebPilot observes the ineffectiveness of interacting with the statictext, adjusts its strategy, and shares this reflection with other nodes. This approach allows WebPilot to dynamically adapt its actions based on new observations, enhancing its decision-making in future scenarios and demonstrating its capacity for adaptive learning. By continuously refining its strategy, WebPilot is able to navigate complex web environments effectively, even when faced with seemingly identical elements that require different interactions. An example of the start page in the GitLab domain is shown in Fig. \ref{actree_example}.

\subsubsection{Long-Horizon Task Complexity}
Tasks in WebArena often require long-horizon planning, involving an extensive sequence of actions that challenge the ability of agents to maintain effective reasoning throughout the process. WebPilot overcomes this challenge by decomposing complex tasks into smaller, more manageable subtasks, each focused on a specific, simpler goal, and by dynamically adapting its plans based on completed subtasks and new observations. For instance, when a task requires inviting a user to a specific project, as shown in Fig. \ref{481_traj}, the optimal path involves navigating to the target project, entering the member management pages, and performing the invitation process. \textit{Planner} of WebPilot divides these steps into individual subtasks, which are easier to solve sequentially. Without this decomposition, the agent might struggle with the complexity of the main task. For example, as shown in Fig. \ref{481_traj}, SteP can become stuck repeatedly clicking the "Settings" button because this action is not adequately covered in its predefined state-specific action policy. Additionally, long-horizon tasks, which require extended action sequences, are prone to deviations from the optimal path due to execution errors. WebPilot addresses this by using \textit{Controller} to ensure each subtask is properly completed, and by treating these subtasks as checkpoints. This allows WebPilot to revert to a previous state if necessary, avoiding the need to re-execute lengthy tasks and thereby maintaining both efficiency and accuracy.

\subsection{Characteristics of MiniWoB++}
\begin{figure*}
    \centering
    \begin{tabular}{c c c}
    \begin{minipage}[b]{0.2\textwidth}
    \centering
    \raisebox{-.1\height}{\includegraphics[width=\linewidth]{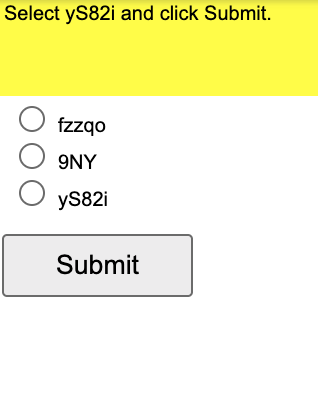}}
    \end{minipage}
    &
    \begin{minipage}[b]{0.2\textwidth}
    \centering
    \raisebox{-.1\height}{\includegraphics[width=\linewidth]{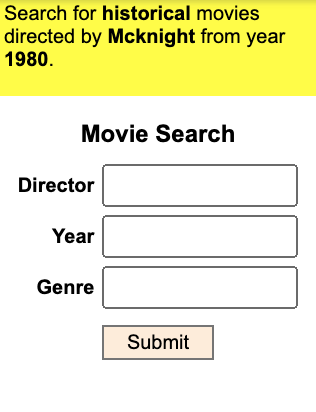}}
    \end{minipage}
    &
    \begin{minipage}[b]{0.2\textwidth}
    \centering
    \raisebox{-.1\height}{\includegraphics[width=\linewidth]{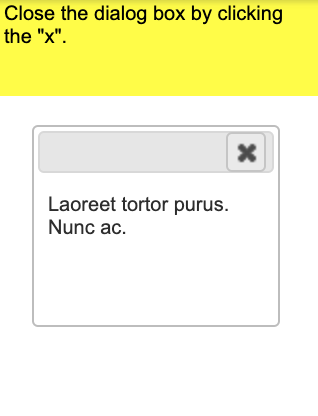}}
    \end{minipage}
    
    \end{tabular}
    \caption{MiniWoB++ Examples.}
    \label{miniwob_example}
\end{figure*}

MiniWoB++ is an environment specifically designed to evaluate the performance of web agents across a range of simpler yet diverse web tasks. We use this environment to test the applicability of WebPilot in straightforward web tasks and to assess its performance across different environments. However, the simplicity of MiniWoB++ limits the potential for significant improvements with WebPilot. As shown in Fig. \ref{miniwob_example}, the tasks are strongly instruction-driven and closely aligned with specific actions, making it easier for agents to identify and execute the correct actions without requiring advanced reasoning or strategic planning. In many cases, these tasks demand only one or two actions to achieve the desired outcome, further reducing their complexity. The environment in MiniWoB++ is intentionally designed with few elements, as shown in Fig. \ref{fig:movie_search}, simplifying decision-making and minimizing the need for complex reasoning. As a result, the advanced capabilities of WebPilot, which are optimized for more complex, dynamic environments, do not offer a substantial advantage in MiniWoB++. Consequently, the performance gains achieved by WebPilot in MiniWoB++ are less pronounced, as these tasks do not necessitate the sophisticated strategies required in more challenging scenarios like those found in WebArena.

\begin{figure}[htbp]
    \centering
    \begin{framed}
    \begin{verbatim}
    [1] <body> 
        [3] <div> id: 'area' 
                [4] <div> text: 'Movie Search' classes: 'title-div'
                [5] <table> 
                        [6] <tbody> 
                                [7] <tr> 
                                        [8] <th> text: 'Director' 
                                        [9] <td> 
                                                [10] <input_text> 
                                [11] <tr> 
                                        [12] <th> text: 'Year' 
                                        [13] <td> 
                                                [14] <input_text> 
                                [15] <tr> 
                                        [16] <th> text: 'Genre' 
                                        [17] <td> 
                                                [18] <input_text> 
                [19] <div> text: 'Submit' classes: 'final'
    \end{verbatim}
    \end{framed}
    \caption{An example of MiniWoB++ DOM elements.}
    \label{fig:movie_search}
\end{figure}

\subsection{Implementation Details}
For both WebArena and MiniWoB++, the implementation is configured with a maximum node count of 10, an exploration bias of 5, a search depth limit of 5, and 3 branches. These hyperparameters are selected to balance exploration and efficiency, as demonstrated in the illustrative example provided in §\ref{Ablation_Hyper}.

\section{Additional Illustrative Examples from Ablation Studies}

\subsection{Illustrative Example: Impact of Hyperparameters on Performance}
\label{Ablation_Hyper}
\subsubsection{Max Node Count $n_{\text{max}}$} Based on our experiments and considering the task complexity in WebArena, we found that setting the number of max node counts to 10 is optimal. While increasing the number of max node counts can slightly improve success rates for particularly challenging tasks, it often leads to wasted resources, making it crucial to find the right balance. This inefficiency arises because many nodes may already be guiding the agent down incorrect paths. Given the tailored design of MCTS in WebPilot, if promising nodes are not found within the first two layers, it typically indicates that the initial paths are flawed. In such cases, it is more effective to use \textit{Controller} to assess task completeness and re-execute the subtask. Conversely, if $n_{\text{max}}$ is set too low, MCTS may struggle to reach the target node. For example, if a subtask requires two to three actions and the branching factor is set to 3, at least two to three layers of search are necessary. With only a few nodes explored, the search may terminate prematurely, missing the correct path.

\subsubsection{Num of Branches}
In MCTS, the number of branches refers to the maximum number of children that a node can expand into. In classical MCTS algorithms, such as those used for board games, this parameter is often set very high. However, in WebPilot, we prioritize the first few choices informed by the initial intuition of LLMs, which are generally more valuable. Therefore, we limit the number of branches. Given the complexity of the WebArena environment, as discussed in §\ref{Challenges in WebArena}, it remains crucial to maintain a sufficient number of branches to ensure adequate exploration and accurate decision-making.

Setting the number of branches too high can lead to inefficiencies, as the initial intuition of LLMs tends to provide a more accurate starting point, effectively narrowing the action space. Expanding into too many branches may result in exploring irrelevant nodes. Conversely, if the number of branches is too low, there is a risk of missing the correct path, especially when the initial intuition is off-target. In such cases, the first few explorations are essential for guiding WebPilot in the right direction. These explorations allow the agent to evaluate the nodes linked to the initial intuition, assigning them lower scores through DES to avoid revisiting them, and they generate sibling reflections that refine subsequent actions and improve decision-making.

Through extensive testing, we found that the initial intuition of LLMs is often accurate, and the number of relevant elements for a given subtask is usually no more than three. For example, when WebPilot is tasked with navigating to the "Members" page from the homepage of a specific repository, the relevant elements are typically "Project Information," "Repository," and "Settings." Other elements are excluded by LLMs. Therefore, we set the number of branches to 3.

\subsubsection{Exploration Bias $w_{puct}$}
The Exploration Bias plays a critical role in shaping the exploration process within the MCTS framework. Adjusting the Exploration Bias can shift the exploration strategy more toward either a depth-first or breadth-first search. In the GitLab domain, for instance, the agent needs to navigate to the "Members" page of a specific project. The link to this page is not directly visible, requiring the agent to explore the webpage thoroughly. Once the dropdown menu containing the target link is opened, this node should be assigned a high value \(S_{\text{total}}\) due to its future promise \(S_{\text{fut}}(o_{t+1})\). However, if the exploration bias is set too high, the agent may be less inclined to move toward this promising node, potentially causing inefficiencies. Conversely, an appropriately set exploration bias allows \textit{Explorer} to efficiently direct the search toward the optimal path. 

\subsection{Illustrative Example: Impact of Reward Function}
Our approach introduces a more nuanced reward function, consisting of two key components: action effectiveness $S_{\text{eff}}(a_t)$ and the future promise $S_{\text{fut}}(o_{t+1})$ of the current observation.

\begin{wrapfigure}{r}{0.35\linewidth}
    \centering
    \vspace{-10pt}
    \includegraphics[width=\linewidth,height=4.5cm]{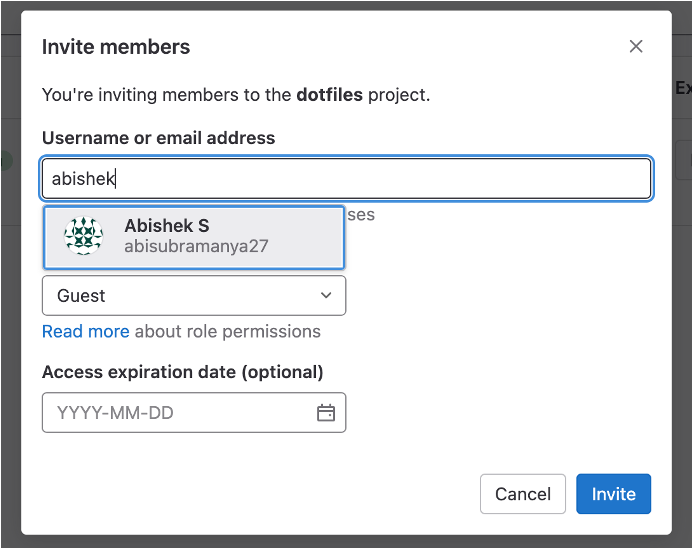}
    \caption{Example of Value function.}
    \label{value function}
\end{wrapfigure}
Action effectiveness $S_{\text{eff}}(a_t)$ assesses how effectively an action contributes to the completion of a subtask. It assigns high scores to actions that are essential for achieving the goal of the subtask, such as "bringing up a dropdown menu" or "choosing the proper sorting option" when the subtask is "sorting items in a specific order". Conversely, it penalizes actions that are irrelevant or do not contribute to the objective. For instance, if the goal of the subtask is to navigate to the homepage of an account in the shopping domain, clicking on "Goods" would not be beneficial and would receive a low action effectiveness score. This evaluation helps avoid unnecessary actions and directs WebPilot more efficiently toward its target.

Future promise $S_{\text{fut}}(o_{t+1})$ is crucial for guiding the exploration of the agent by indicating whether the current page can lead to the target page. For instance, if the agent needs to navigate to a specific page and a link to it is visible in the current state, the evaluator will assign a higher value to this state, encouraging the agent to explore this page further.

A more detailed example can illustrate how these components work together in practice. As illustrated in Fig. \ref{value function}, consider a subtask where the objective is to invite Abishek as a project member. After WebPilot enters Abishek's name into the textbox, the action is considered highly effective because it directly addresses a necessary requirement of the subtask. Simultaneously, the current observation shows the target user in the dropdown menu, indicating a state with high future promise, as the next crucial step is to select the target user from the dropdown menu.

\begin{tcolorbox}[colback=white, colframe=black, title=\texttt{<executed\_action\_effectiveness>: 9}, width=\linewidth, halign=left]
\begin{lstlisting}[language=TeX, basicstyle=\ttfamily\footnotesize, numbers=none, xleftmargin=0pt, frame=none]
The executed action is effective as it directly addresses the task requirement and aligns with the goal of adding Abishek's username to the invitation dialog. It is a necessary step in the process.
\end{lstlisting}
\end{tcolorbox}

\begin{tcolorbox}[colback=white, colframe=black, title=\texttt{<future\_promise>: 8}, width=\linewidth, halign=left]
\begin{lstlisting}[language=TeX, basicstyle=\ttfamily\footnotesize, numbers=none, xleftmargin=0pt, frame=none]
The current state of the webpage, with Abishek's username already filled in the invitation dialog, indicates progress toward the goal. The necessary action has been completed, and the next steps involve selecting the correct username from the list, which seems feasible based on the available information.\end{lstlisting}
\end{tcolorbox}

\subsection{Illustrative Example: Impact of MVB}
\begin{figure}
    \centering
    \includegraphics[width=1\linewidth]{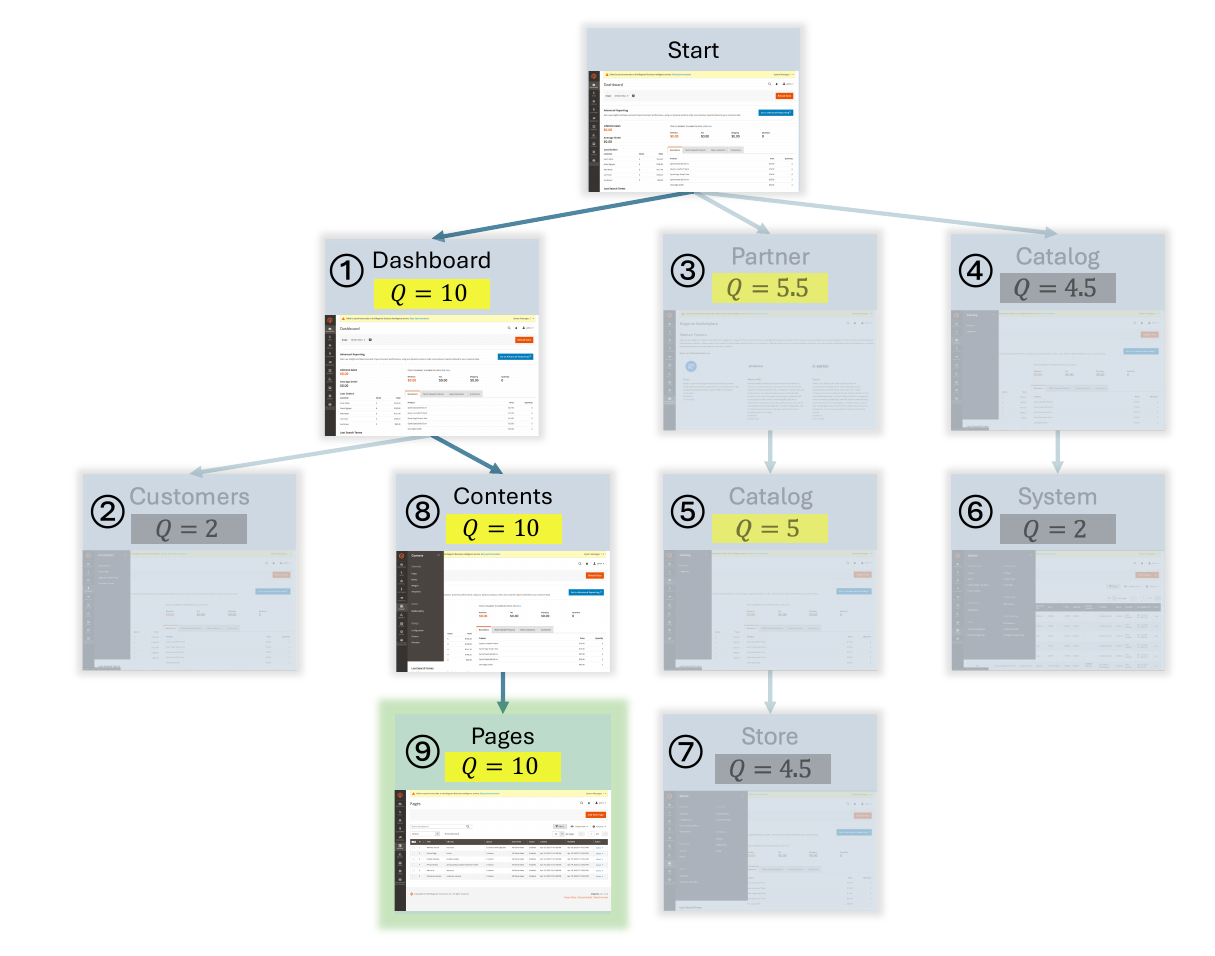}
    \caption{Impact of Maximal Value Backpropagation (MVB) during the Local Optimization stage of WebPilot, illustrating its role in accelerating convergence to target states by prioritizing high-value paths. This figure provides a detailed view of the execution process, with Fig. \ref{fig:mcts_4_stages} in the main paper, offering a simplified overview. The subtask is "Navigate to the 'Pages' site". The index numbers beside the screenshots indicate the exploration order within this search process.}
    \label{q_update}
\end{figure}

If Maximal Value Backpropagation (MVB) is not utilized and a simple averaging method is employed for backpropagation instead, the tree search will reach the target state more slowly and may even fail under a limited max node count \(n_{\max}\). Consider Fig. \ref{q_update} as an example, where the goal is to navigate to the "Pages" section of a website. If the “Contents” section is expanded and the “Pages” link becomes visible, leaving only one step to reach the target, this state would be highly rated. However, without MVB, this higher score would be partially diluted as it propagates back to the parent node, potentially causing the search to deviate from this promising path in subsequent explorations, ultimately delaying the achievement of the target. This approach is particularly crucial for tasks involving website interactions, especially when certain target states are concealed and need to be uncovered. However, information-seeking tasks also benefit from this design, as many subtasks require exploring the web interface before extracting key information.

\subsection{Illustrative Examples of Ablation Study Findings}
Here, we use Fig. \ref{481_setting} to illustrate key findings from the ablation study in the main paper, highlighting the impact of specific components on the performance of WebPilot.

\subsubsection{Illustrative Example: Importance of Child Reflection \(\mathcal{R}_{c_t}\) for Maintaining Coherent Thought Processes}
The child reflection mechanism is essential for maintaining a coherent thought process during exploration, as it aligns with the principles of Chain-of-Thought reasoning. To illustrate, consider the example in Fig. \ref{481_setting}, where the subtask is to "invite Abishek". This subtask involves four sequential actions: bringing up the invite dialog, typing Abishek's name, selecting the target user from the dropdown menu, and completing the invitation. The coherence of these actions is critical. For instance, when the invite dialog is brought up, due to the way WebArena extracts the actree, elements outside the dialog remain visible. Interacting with any of these other elements would close the dialog, disrupting the process. Similar to selecting the target user from the dropdown menu. After WebPilot completes the action of typing the username, the child reflection mechanism suggests the next logical step should be selecting the corresponding menu item to complete the invitation.

\begin{tcolorbox}[colback=white, colframe=black, title=\texttt{<reflection\_for\_child>}, width=\linewidth, halign=left]
\begin{lstlisting}[language=TeX, basicstyle=\ttfamily\footnotesize, numbers=none, xleftmargin=0pt, frame=none]
Since the task aims to invite Abishek, and "Executed Actions" shows a finished action of typing "Abishek" in the search box, and a dropdown menu including "Abishek" appears, the next action should be selecting the target username to ensure inviting him.
\end{lstlisting}
\end{tcolorbox}

This reflection ensures continuity and helps the model generate accurate and contextually relevant actions, preserving the logical flow necessary for complex decision-making. Without child reflection, this coherence would be lost, leading to significant performance declines, particularly in tasks where maintaining a consistent thought process is critical.

\subsubsection{Illustrative Example: Critical Role of Sibling Reflection \(\mathcal{R}_{s_t}\) in Effective and Diverse Exploration}
Sibling reflection plays a crucial role in optimizing exploration within MCTS, especially in complex tasks where diverse and effective exploration is needed. This mechanism allows WebPilot to provide sibling nodes with failure experiences from previous attempts, which is particularly valuable since these sibling nodes are generated under identical conditions. By sharing insights from explored paths, sibling reflection helps the agent avoid redundant actions and focus on more promising alternatives.

For instance, in Fig. \ref{481_setting}, during the subtask "Navigate to Member's Page," WebPilot initially struggles to locate the correct link to the "Members" page. WebPilot chooses to click on "Repository" in an attempt to find the members page. However, upon realizing that this action led to the "Files" page instead of the desired "Members" page, WebPilot generates the following sibling reflection to guide the subsequent action toward the correct path:

\begin{tcolorbox}[colback=white, colframe=black, title=\texttt{<reflection\_for\_sib>}, width=\linewidth, halign=left]
\begin{lstlisting}[language=TeX, basicstyle=\ttfamily\footnotesize, numbers=none, xleftmargin=0pt, frame=none]
The "Observation Description" does not align with the "Expectation" as the action led to the "Files" page instead of the "Members" page. The "action_intent" is appropriate for accessing the members page of the "dotfiles" repository through the sidebar. To correct this, the next action should be to click on the "Members" link in the sidebar to navigate to the members page of the "dotfiles" repository.
\end{lstlisting}
\end{tcolorbox}

This reflection allows the agent to adjust its strategy in real time, redirecting its focus to the correct link in the sidebar. By doing so, WebPilot enhances exploration effectiveness, ensuring that it pursues high-potential paths and avoids repeating errors. This mechanism is particularly impactful in web interaction tasks, as it enables WebPilot to efficiently navigate through complex environments by learning from its previous attempts.

\subsubsection{Illustrative Example: Significance of \textit{Controller} in Ensuring Subtask Accuracy and High-Level Decision-Making}
\textit{Controller} plays a critical role in WebPilot by determining when to stop a subtask and assessing its completeness \(\text{Comp}_i\), upon which it generates the subtask reflection \(\mathcal{R}_{\text{sub}_i} \). Unlike classical MCTS methods, which focus on selecting the next action within limited computational resources in a single search process, the goal of WebPilot is to reach the terminal node that satisfies the subtask requirements. Therefore, \textit{Controller} is essential for deciding whether the target node has been reached during the search, efficiently guiding WebPilot by ending the search when the target is achieved.

Subtask Reflection is generated by \textit{Controller} after evaluating the execution of the current subtask \(\mathcal{T}_i\). \textit{Controller} is responsible not only for determining when to stop a subtask but also for extracting critical insights from any failed attempts. For instance, in a task requiring the agent to click on "All" tabs, the first action might correctly click the appropriate tab, but a subsequent action might mistakenly navigate to an incorrect page. The task is also flagged for termination because the necessary action of clicking the link "All" is executed. When \textit{Controller} analyzes the completeness of the current subtask, it identifies this error and generates a Subtask Reflection, suggesting that "clicking 'All' tabs" should be the final action in similar tasks. This reflection is then used to guide the re-execution of the subtask, ensuring that the agent avoids previous mistakes and completes the task accurately. Following is an example of the subtask reflection generated by WebPilot when it is tasked with navigating to the “Orders” page in the CMS domain but failed on the first attempt:

\begin{tcolorbox}[colback=white, colframe=black, title=\texttt{<subtask\_reflection>}, width=\linewidth, halign=left]
\begin{lstlisting}[language=TeX, basicstyle=\ttfamily\footnotesize, numbers=none, xleftmargin=0pt, frame=none]
To successfully complete the task, focus on directly navigating to the "Order Report" page instead of getting sidetracked by other sections like "Marketing" or "Dashboard". Pay close attention to the menu options and sub-options to reach the target page efficiently.
\end{lstlisting}
\end{tcolorbox}

Furthermore, \textit{Controller} assesses the completeness of the current subtask to determine if the plan requires updating, and then provides the necessary recommendations/reflection to \textit{Planner} for plan refinement. For instance, in a task where the objective is to "Check out the merge requests assigned to me," \textit{Planner} initially generates a plan to navigate to the "All merge requests" page and then apply a filter for "assigned to me." However, during the execution of the first subtask, \textit{Explorer} discovers a direct link to the "merge requests assigned to me" page on the main interface, effectively completing the task ahead of schedule. \textit{Controller} then reassesses the plan, determining that the remaining subtasks are no longer necessary, leading to an update of the plan. Following is the completeness assessment for this task generated by the \textit{Controller}: 

\begin{tcolorbox}[colback=white, colframe=black, title=\texttt{<task\_completeness>}, width=\linewidth, halign=left]
\begin{lstlisting}[language=TeX, basicstyle=\ttfamily\footnotesize, numbers=none, xleftmargin=0pt, frame=none]
The main task of navigating to the All merge requests through the top bar is completed successfully as the webpage displays merge requests assigned to "Byte Blaze".
\end{lstlisting}
\end{tcolorbox}

\subsubsection{Illustrative Example:Essential Role of \textit{Planner} in Enhancing Performance in Complex Tasks}
\textit{Planner} is crucial for breaking down complex tasks into simpler, focused subtasks, each with a clear goal. This approach prevents the agent from getting stuck in repetitive actions or being distracted by irrelevant information. In the task shown in Fig. \ref{481_setting}, where the goal is to "invite Abishek to the 'dotfile' repository as a guest," \textit{Planner} of WebPilot effectively decomposes the task. Without this guidance, as seen with SteP, the agent might repeatedly navigate to the "Settings" page, mistakenly assuming it to be the correct path. The \textit{Planner}, however, creates a subtask focused on navigating directly to the "Members" page, allowing the agent to complete the task more efficiently by avoiding unnecessary actions and staying on course.

\section{Illustrative Examples of Agent Behavioral Analysis}
Beyond the example provided in §\ref{behavioral} of the main paper, this section presents additional illustrative examples that demonstrate the behavior of WebPilot to show its effectiveness in mimicking human strategies. 

\subsection{Illustrative Example: WebPilot Learning Website Functionality by Exploring Unknowns}
The ability of WebPilot to explore unknowns is consistently evident across various tasks. For example, in the CMS domain, a task requires accessing the "Pages" management section. Although the link is not immediately visible on the initial dashboard, WebPilot navigates the sidebar to locate it. This exploratory capability is also demonstrated in tasks involving navigation to specific pages via sidebar elements, highlighting the proficiency of WebPilot in adapting to and mastering unfamiliar web environments.

\subsection{Illustrative Example: WebPilot Adapting Strategies Based on New Observations}
The ability of WebPilot to adapt its strategy is further illustrated in the following scenario. When searching for a user profile in the GitLab domain, WebPilot initially anticipates the results to be directly related to users. However, the search results default to displaying information about "Projects," with the "Users" tab hidden. Upon observing the current page layout, WebPilot adapts its action strategy and suggests clicking the "Users" tab in its reflection for subsequent steps. This reflection acts as a hint for future actions, preventing the agent from incorrectly assuming that the user-related results are unavailable based solely on the initial search outcome.

\subsection{Illustrative Example: WebPilot Resolving Ambiguities Through Iterative Refinement}
Ambiguity can be divided into two categories: environmental ambiguity and task ambiguity. In §\ref{behavioral}, we discussed how WebPilot addresses environmental ambiguity. An illustrative example of handling task ambiguity comes from the Reddit domain. The task is to "Post in the most appropriate subreddit and ask for recommendations for noise-canceling headphones." The appropriate subreddit is not immediately apparent, making it challenging to create a comprehensive plan at the outset. To tackle this, \textit{Planner} of WebPilot breaks down the task into a relevant subtask—"Navigate to a relevant subreddit, such as headphones or shopping". \textit{Explorer} then dynamically executes this subtask, selecting the appropriate forum after reviewing the full list of subreddits. Finally, \textit{Controller} validates the execution results to confirm whether the desired subreddit has been reached. This collaborative approach effectively resolves the ambiguity inherent in the task, allowing WebPilot to adaptively navigate complex decision-making scenarios.

\subsection{Illustrative Example: High-Level Demos Generalize Better to Unseen Tasks}
\label{demo_generalize}
High-level demonstrations equip the agent with general web domain knowledge, enabling WebPilot to generate more effective plans, better understand the environment, and even for tasks not explicitly covered in the demonstrations. For example, in the Shopping domain, the \textit{Planner} used these demonstrations, as shown in §\ref{demo} to create a plan for the following task "Find the customer name and email with phone number 2137418080":

\begin{tcolorbox}[colback=white, colframe=black, title=\texttt{<\textit{Planner}>}, width=\linewidth, halign=left]
\begin{lstlisting}[language=TeX, basicstyle=\ttfamily\footnotesize, numbers=none, xleftmargin=0pt, frame=none]
Subtask 1: Navigate to the "Customers" page
Subtask 2: Search for the phone number 2137418080
Subtask 3: Extract the customer's name and email from the search results
\end{lstlisting}
\end{tcolorbox}

Although this specific task is not included in the high-level demonstrations, WebPilot is able to generalize from the provided examples, effectively applying the domain knowledge to generate a suitable plan. This ability to generalize arises from the model focusing on overarching strategies rather than rigid state-action pairs, allowing it to adapt to new tasks by leveraging its understanding of similar tasks. As a result, the \textit{Explorer} successfully completed the task, along with three additional tasks from the same template.

\subsection{Illustrative Example: Strategic Differences Between WebPilot and SteP}
As shown in Tab. \ref{webarena_performance} of the main paper, WebPilot outperforms the current SOTA agent SteP, particularly in the Gitlab domain. Below, we analyze the strategic differences between WebPilot and SteP. The trajectory data is provided by SteP.

\begin{figure}
    \centering
    \includegraphics[width=\linewidth]{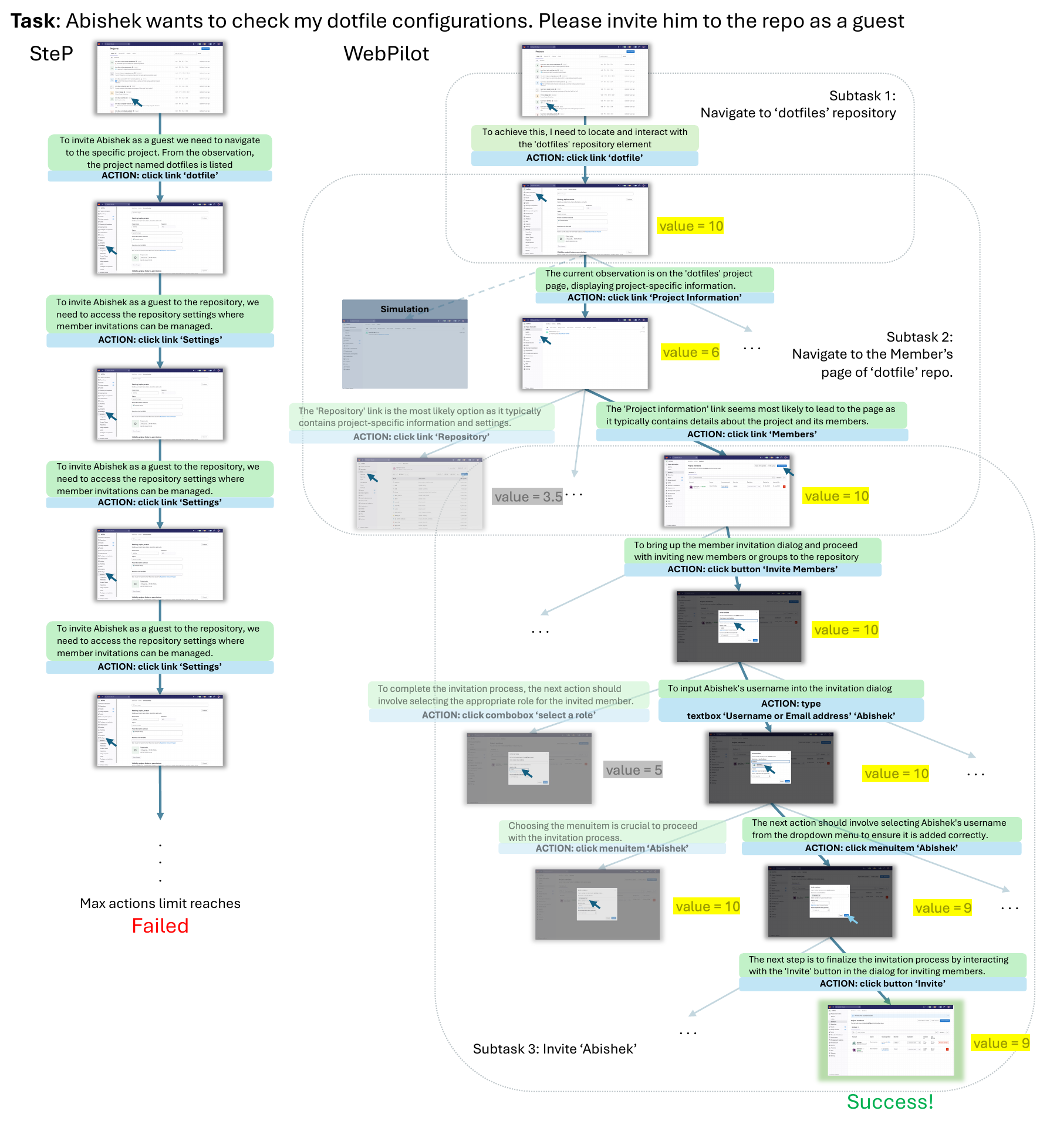}
    \caption{Example trajectories on WebArena for SteP (left) and WebPilot (right). Task: "Invite Members".}
    \label{481_traj}
\end{figure}

As Fig. \ref{481_traj} illustrates, SteP repeatedly attempted to navigate to the "Settings" page of the repository, despite these actions being clearly ineffective. This behavior reveals a critical limitation of SteP: its reliance on predefined action policies, which hinder its ability to adapt when those policies fail to produce the desired outcome. The dependence of SteP on static, state-specific action policies limits its flexibility, causing it to become trapped in a loop of repetitive actions without recognizing the need to backtrack or explore alternative paths. This shortcoming highlights the importance of dynamic reflection, a mechanism that WebPilot employs effectively to avoid such pitfalls.

\begin{figure}
    \centering
    \includegraphics[width=0.45\linewidth]{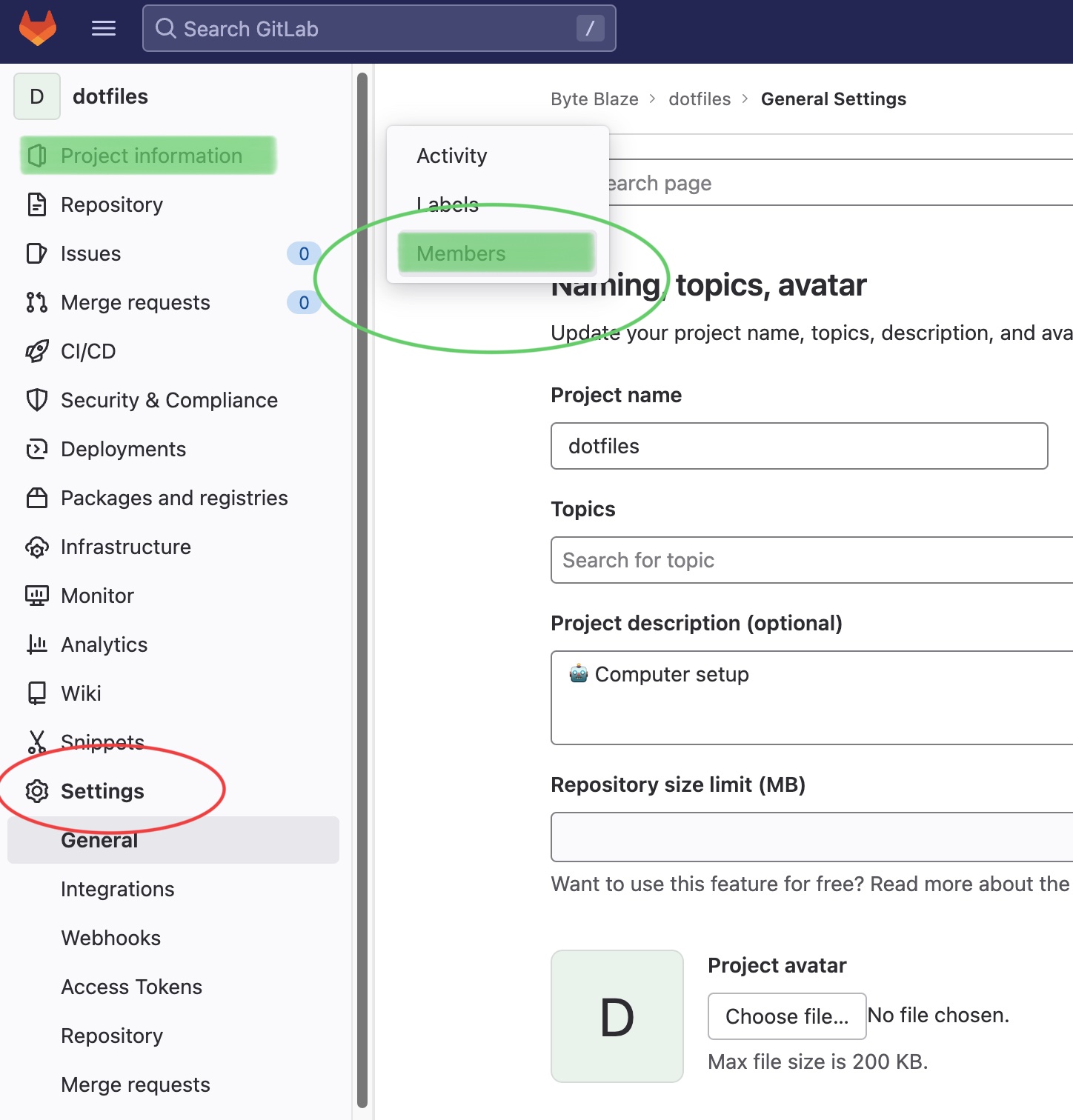}
    \caption{Example of Specific Screenshot: SteP consists that the "setting" may lead to the right page for member management while WebPilot navigates to the "Member" page.}
    \label{481_setting}
\end{figure}

In contrast, WebPilot successfully completed the task by decomposing it into several subtasks using Hierarchical Task Decomposition (HTD), where each subtask focused on a singular, well-defined goal. For instance, when tasked with navigating to the "Members" page, WebPilot avoided distractions by isolating this subtask from the broader task, allowing it to proceed directly to the target. This focused approach not only improved task efficiency but also ensured that WebPilot could adapt to unforeseen challenges by continuously reassessing and refining its strategy in real time. As shown in Fig. \ref{481_setting}, SteP is unable to move beyond the repetitive "Settings" action, ultimately becoming stuck. However, WebPilot leveraged its exploration capabilities to correctly identify and navigate to the appropriate target page, demonstrating the advantage of its flexible, adaptive strategy over the more rigid, policy-driven approach of SteP.

\begin{figure}
    \centering
    \includegraphics[width=1\linewidth]{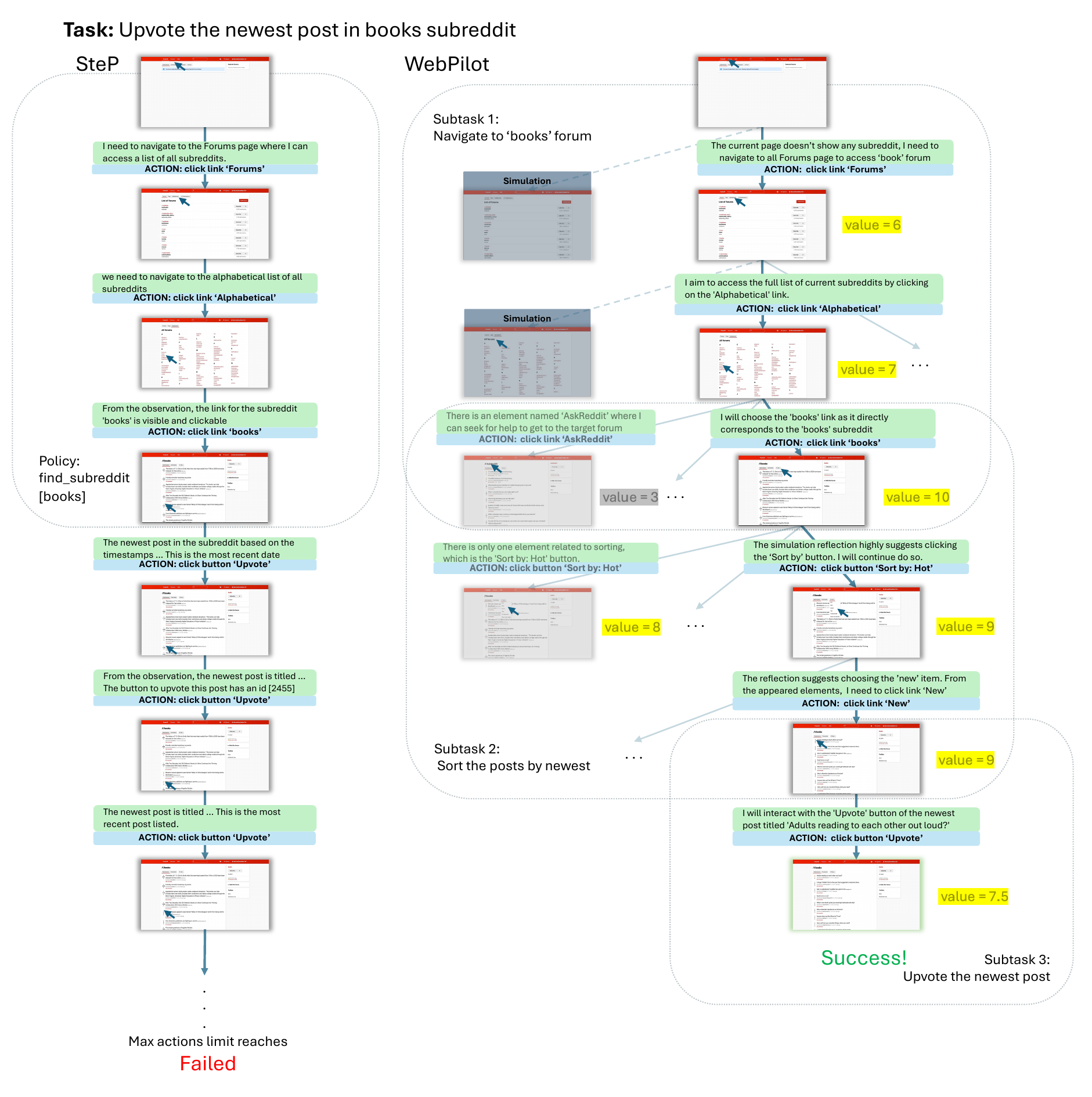}
    \caption{Example trajectories on WebArena for SteP (left) and WebPilot (right). Task: "Upvote the newest post".}
    \label{404_traj}
\end{figure}

Without task decomposition, SteP tends to focus solely on the main task, overlooking important details within it. For example, as shown in Fig. \ref{404_traj}, after reaching a specified forum, SteP repeatedly clicked "UPVOTE" without achieving the intended outcome, ultimately halting its operation. This behavior indicates that when operating outside of its predefined policies, SteP functions like a typical agent, prone to repeating the same errors. In contrast, WebPilot flexibly assesses all aspects of the main task, prioritizes necessary actions, and executes them effectively.

\begin{figure}
    \centering
	\includegraphics[width=1\linewidth]{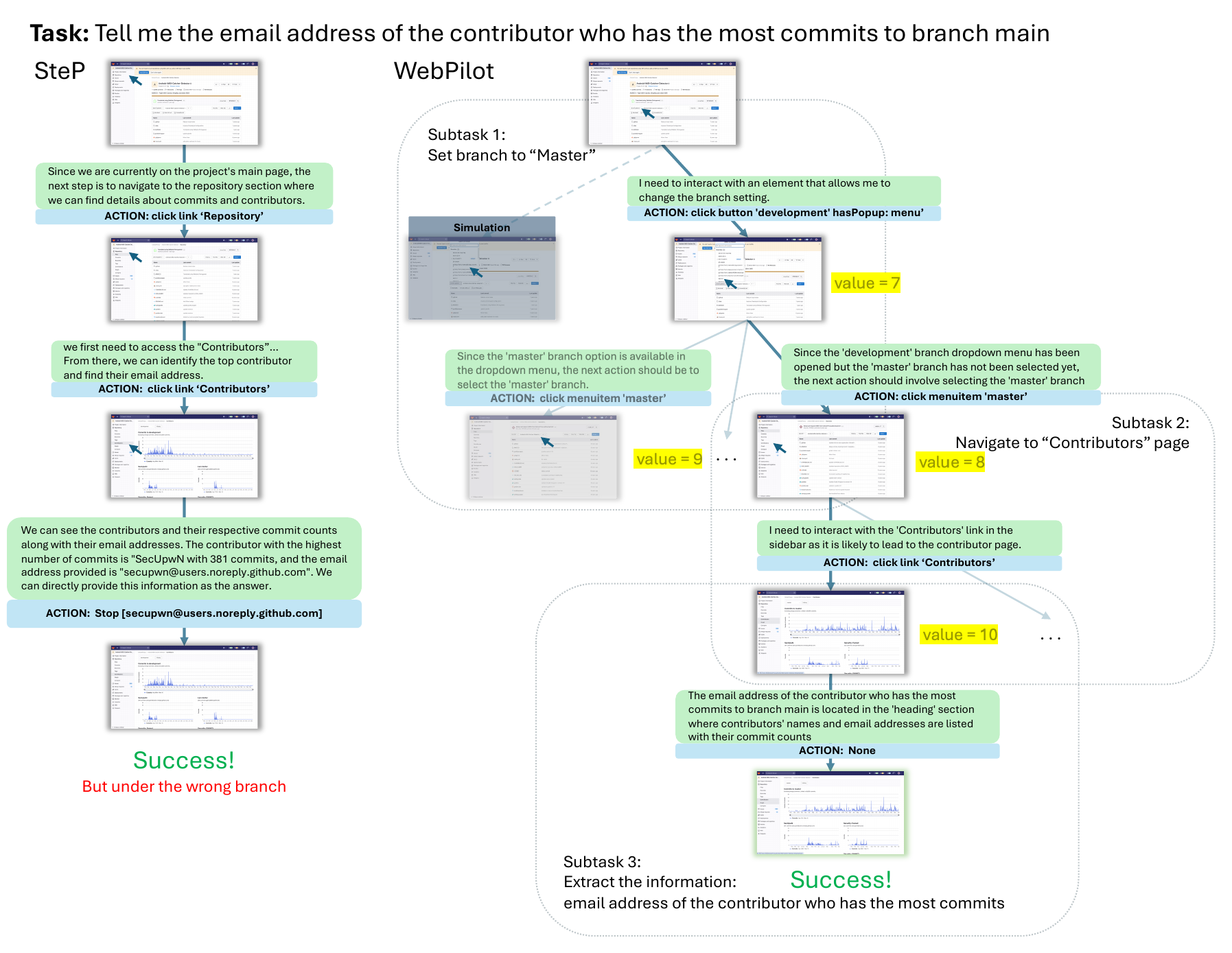}
    \caption{Example trajectories on WebArena for SteP (left) and WebPilot (right). Task: "Extract Contributor email address".}
    \label{784_traj}
\end{figure}

Another example is illustrated in Fig. \ref{784_traj}, where agents are tasked with extracting the information of the contributor who has made the most commits to a specific branch “main” as an IS task. Starting from the repository homepage, SteP navigated directly to the “Contributors” page and immediately provided an answer without correctly setting the branch. Although it received a reward for success, this is purely coincidental—the top contributor of the “development” branch, the default branch, happened to be the same as the top contributor of the “main” branch. This issue reflects a common problem observed in realistic web environments, such as those in WebArena \cite{zhou2023webarena}, where GPT-4 agents often latch onto the first piece of related information they encounter without fully verifying its relevance or accuracy. However, WebPilot addresses this challenge by utilizing HTD to decompose tasks systematically, ensuring no critical details are overlooked. For instance, the first subtask, “set the branch to ‘main’,” guaranteed that the information provided met the task requirements accurately.

\section{More Discussions on Limitations}
\subsection{Comparative Analysis of Agent Behavior with GPT-3.5 and GPT-4}
\subsubsection{Case 1: Environment Understanding}
Compared to GPT-3.5, GPT-4 demonstrates a significantly more accurate understanding of environments. This is particularly evident when dealing with pages containing items with similar attributes. For instance, on a page with multiple orders, each having its own status—such as "Cancelled," "Complete," and "Pending"— GPT-4 is notably more effective at selecting a specific order based on a given attribute value. A detailed example of this can be found in \ref{limitation_text}.

\subsubsection{Case 2: Utilizing the "Scroll" Action to Explore}
We observe that WebPilot, when powered by GPT-4, is more adept at exploring environments using the "Scroll" action. A notable example is in the task of modifying a user's homepage, where the target textbox only becomes visible after scrolling down within the profile settings page. While WebPilot with GPT-3.5 struggles with this operation and fails to execute it successfully, GPT-4 handles it with greater effectiveness.

\subsubsection{Case 3: Plan Generation for Unseen Tasks}
While high-level demonstrations are beneficial for generating plans for unseen tasks, we observe that WebPilot with GPT-4 still outperforms GPT-3.5 in planning. This is evidenced by the generally higher success rates. Notably, in certain task templates where GPT-3.5 struggles to generate effective plans, GPT-4 demonstrates significantly improved results.

\subsubsection{Case 4: Knowledge Precision}
In the Map domain, for the task "Show me the path and travel time from the Big Apple to the biggest city in Maine", we encourage \textit{Planner} to leverage its built-in knowledge—specifically the knowledge embedded within the LLM—to generate a plan for searching the route between "New York" and "Portland". While WebPilot with GPT-4 successfully generates the correct plan, WebPilot with GPT-3.5 struggles to output the correct city names.

\subsection{Text-based Observation vs. Vision-based Observation}
\label{limitation_text}
\begin{figure}
    \centering
    \includegraphics[width=1\textwidth]{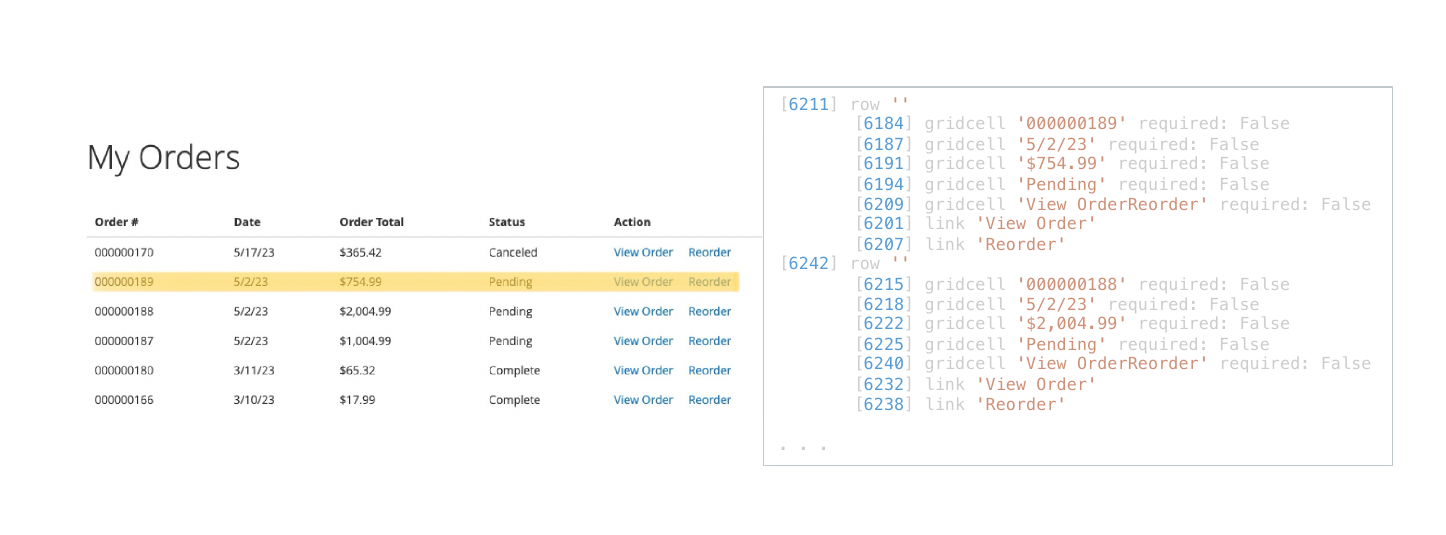}
    \caption{Example of an order number recognition task: The agent must identify and provide the order number of the most recent pending order, in this case, 000000189.}
    \label{232_order}
\end{figure}

When processing text-based web observations, such as actrees, LLMs often interpret page content based on proximity principles. For instance, when asked to identify the order number of the most recent pending order, the correct answer should have been 189, as shown in Fig. \ref{232_order}. However, due to the layout of the text—where a different element is positioned closer—the LLM incorrectly identified the order number, resulting in an erroneous interpretation.

The challenge with text-based comprehension arises from the absence of visual context, which limits the ability of the agent to accurately interpret spatial relationships and distinguish between textually similar elements. Unlike humans, who can easily parse a webpage by leveraging visual cues such as color, layout, and spacing, LLMs rely solely on textual information. This reliance on text makes it difficult for the model to discern the correct relationships between elements, especially when similar or related items are positioned close to each other, increasing the likelihood of misinterpretation.

\begin{figure}[H]
    \centering
    \includegraphics[width=0.95\linewidth]{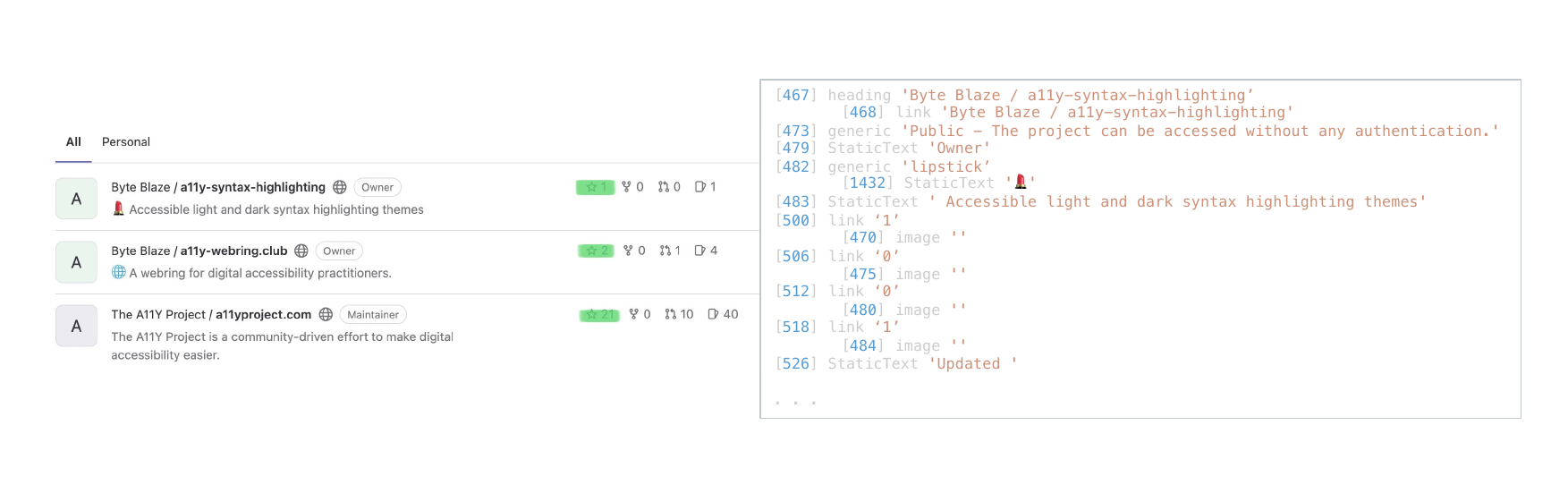}
    \caption{Screenshot and actree representation of the star information in the task "Tell me the full names of the repositories where I made contributions and they got more than 100 stars".}
    \label{168}
\end{figure}

Additionally, certain visual information cannot be adequately represented through actrees, further complicating the ability of LLM to accurately interpret web content. For example, in the task "Tell me the full names of the repositories where I made contributions and they got more than 100 stars", the star information is presented solely as a link with a numeric value and an "image" item in the actree, as shown in Fig. \ref{168}. The lack of visual representation makes it impossible for the LLM to correctly extract and interpret this information.

\subsection{Dataset Limitation}
\label{limitation_dataset}
\subsubsection{Element Not Directly Observable}
\begin{figure}[H]
    \centering
    \includegraphics[width=0.95\linewidth]{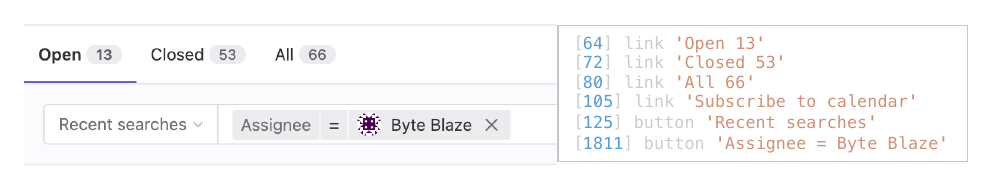}
    \caption{Example of unchanged elements: "Open," "Closed," and "All" tabs on the issue inspection page in WebArena.}
    \label{173_open_close}
\end{figure}

On certain pages of WebArena, the state changes of specific elements are not effectively conveyed through the actree. For example, on the issue inspection page, clicking the "Open," "Closed," or "All" tabs triggers different filters that display the status of various issues. However, despite significant changes in the content displayed, the text-based state of the tab elements themselves remains unchanged. As a result, the agent, which relies on comparing the state of the target element before and after execution, may incorrectly interpret the action as a failure. Fig. \ref{173_open_close} shows the possible issue filtering tabs and their text-based representation, with no additional annotations to indicate their states.

\subsubsection{Invisible Dropdowns}
In several domains within WebArena, there have been instances where certain dropdown menus fail to expand, limiting the performance of WebPilot in specific tasks.
\section{Algorithm Details}

\subsection{Summary of the Function of Each Agent}

\begin{table}[H]
\centering
\begin{tabular}{r|c}
\hline
Agent  & Functions            \\
\hline
\textit{Planner}                        & Plan Generation; Plan Refinement \\
\textit{Controller}                     & Subtask Termination Judgment; Subtask Completion Assessment; Strategic Reflection Generation \\
\textit{Extractor}                      & Information Extraction \\
\textit{Explorer}                       & Action Generation; Observation Analysis; Tactical Reflection Generation \\
\textit{Apprasier}                      & State Assessment \\
\textit{Verifier}                       & Action Formatting; Action Deduplication \\ \hline
\end{tabular}
\caption{Summary of the Function of Each Agent.}
\label{agent_func}
\end{table}

\subsection{Detailed Description of Action Effects $\text{Effect}(a_t)$ in RENE} After executing an action, \textit{Explorer} assesses the resulting changes in the environment, referred to as the action effect $\text{Effect}(a_t)$. This effect is then compared with the intended outcome $\mathcal{I}_t$ by \textit{Appraiser} to determine how effectively the action has achieved its goal.

The assessment process begins by checking whether the base URLs of the previous and current web pages are identical. If the base URLs differ, it indicates that the agent has navigated to a new page. In such cases, WebPilot focuses on describing the new web page to ascertain the current location of the agent. Conversely, if the base URLs remain unchanged, it suggests that only elements on the current page have been modified. In this scenario, \textit{Explorer} prompts the agent to describe these minor changes, such as the opening of a dropdown menu. WebPilot then specifies the location of the dropdown and identifies any new elements that have appeared, which are likely to be relevant for subsequent interactions. This process ensures that WebPilot remains responsive and context-aware in its exploration.

However, it is important to note that in dialogue scenarios, subtle differences in natural language expressions often have less impact on the overall conversation. Consequently, models may tend to overlook minor variations in their observations, a challenge highlighted in WebArena \cite{zhou2023webarena}. These distinct prompts—"What kind of page is reached?" for new pages and "What elements have changed?" for modifications within the same page—guide \textit{Explorer} in accurately identifying and articulating the specific changes that have occurred, thereby enhancing the ability of WebPilot to navigate dynamic environments effectively.

\subsection{High-level Demonstrations for the CMS Domain}
\label{demo}
\begin{framed}
\begin{verbatim}
Example 1:
    **Main Task**: 'Generate a tax report for 2022 Q3.'
    **Decomposed Plan**:
    [
    	{REASONING PROCESS}
     'subtask': 'Bring up the 'REPORTS' section',
     {REASONING PROCESS}
     'subtask': 'Navigate to 'Tax Report' page',
     {REASONING PROCESS}
     'subtask': 'Set date range from 7/1/2022 to 30/9/2022 and generate the 
     report. ',
    ]
Example 2:
    **Main Task**: 'Edit the product 'Nike Flyknit running shoes', set its weight 
    to 0.3 kg and reduce the price by $10.'
    **Decomposed Plan**:
    [
    	{REASONING PROCESS}
     'subtask': 'Navigate to the 'Products' page',
     {REASONING PROCESS}
     'subtask': 'Search for the product 'Nike Flyknit running shoes' and access 
     its detailed page',
     {REASONING PROCESS}
     'subtask': 'Set the weight of the product to 0.3 kg',
     {REASONING PROCESS}
      'subtask': 'Set the price of the product to $99',
    ] 
Example 3:
    **Main Task**: 'Find order 221 and send the confirmation email to the 
    customer'
    **Decomposed Plan**:
    [
     {REASONING PROCESS}
     'subtask':  'Navigate to the 'Orders' page',
     {REASONING PROCESS}
     'subtask': 'Search for order ID 000000221. ',
     {REASONING PROCESS}
     'subtask': 'Navigate to the detail view page of the order 000000302',
     {REASONING PROCESS}
     'subtask': 'Send the confirmation email to the customer',
    ]
Example 4:
    **Main Task**: 'Edit the page 'Order Received' to 'Thanks for choosing us'. '
    **Decomposed Plan**:
    [
     {REASONING PROCESS}
     'subtask': 'Navigate to the 'Pages' section',
     {REASONING PROCESS}
     'subtask': 'Navigate to the edit page of the 'Order Received' page',
     {REASONING PROCESS}
     'subtask': 'Change the title 'Order Received' to 'Thanks for choosing us'. ',  
    ]
    \end{verbatim}
\end{framed}

\subsection{Pseudocode for WebPilot}

\begin{algorithm}
\label{algo}
\caption{WebPilot}
\begin{algorithmic}[1]
\REQUIRE task $\mathcal{T}$; max node count $n_{\text{max}}$; maximum scroll time $n_{\text{scroll}_{\text{max}}}$; environment $\mathcal{E}$; transition function $\mathcal{F}$; time $t = 0$; exploration bias $w_{puct}$; state $s_{\text{initial}} $; observation $ o_{\text{initial}}$; action $a_t = \emptyset$; plan $\mathcal{P} = \emptyset$; finished subtasks $\mathcal{P}' = \emptyset$ history $\mathcal{H} = \emptyset$; action spaces $ A$

\STATE $ s_0 \xleftarrow{} s_{\text{initial}}, o_0 \xleftarrow{} o_{\text{initial}} $
\STATE $ \{\mathcal{T}_1, \mathcal{T}_2, \ldots, \mathcal{T}_n\} \xleftarrow{} \text{\textit{Planner}}(\mathcal{T}, o_{\text{initial}}, \text{demonstrations}) $ \hfill $\triangleright$ \textbf{HTD}
\STATE $ \mathcal{P} \xleftarrow{} \{\mathcal{T}_1, \mathcal{T}_2, \ldots, \mathcal{T}_n\} $

\WHILE{$ \mathcal{P} \neq \emptyset$}
    \STATE $ \mathcal{T}_i \xleftarrow{} \mathcal{P}.\text{pop}(0) $
	\STATE $ n_{\text{scroll}} \xleftarrow{} 0 $
    \IF{ $ \mathcal{T}_i $ involves information extraction }
        \WHILE{$ n < n_{\text{scroll}_{\text{max}}}$}
            \IF{$ \text{Answer} \xleftarrow{} \textit{Extractor}(\mathcal{T}_i, o_t) $}
                \STATE break
            \ELSIF{$ a_{\text{scroll}} \xleftarrow{} \textit{Extractor}(\mathcal{T}_i, o_t) $}
                \STATE $ s_{t+1}, o_{t+1} \xleftarrow{} \mathcal{F}(a_{\text{scroll}}, s_t) $
                \STATE $ t \xleftarrow{} t+1 $
            \ENDIF
            \STATE $ n \xleftarrow{}  n+1 $
        \ENDWHILE
    \ELSE
        \WHILE{$ n < n_{\text{max}} $}
           \STATE $ n \xleftarrow{}  n+1 $
			\STATE $ t \xleftarrow{} 0, \mathcal{H} \xleftarrow{} \emptyset $
            \WHILE {$ N(s_t) > 0 $}
        	    \STATE $ N(s_t) \xleftarrow{} N(s_t) + 1 $ \hfill $\triangleright$ \textbf{GOS}
	        	\STATE $ a_t \xleftarrow{} \arg \max_{a \in A_t} \left[Q(s_t,a) + w_{puct} \frac{\sqrt{\sum_b N(s_t,b)}}{1+N(s_t, a)} \right] $ 
	        	\STATE $ \mathcal{H}_{t+1} \xleftarrow{} [\mathcal{H}_t, a_t] $
	        	\STATE $ s_{t+1}, o_{t+1} \xleftarrow{} \mathcal{F}(a_t, s_t) $
	        	\STATE $ t \xleftarrow{} t+1 $    	
            \ENDWHILE
            
            \STATE $ a_t, \mathcal{I}_t \xleftarrow{} \textit{Explorer}(o_t, \mathcal{T}_i, H_t, \{\mathcal{R}_{\text{sim}_t}, \mathcal{R}_{p_t}, \mathcal{R}_{s_t}, \mathcal{R}_{\text{sub}}\}, \mathcal{C}_{t-1}) $
            	\hfill $\triangleright$ \textbf{RENE}
        	\STATE $ \mathcal{H}_{t+1} \xleftarrow{} [\mathcal{H}_t, a_t] $
            \STATE $ s_{t+1}, o_{t+1} \xleftarrow{} \mathcal{F}(a_t, s_t) $
			\STATE $ \text{Effect}(a_t) \xleftarrow{} \textit{Explorer}(o_{t+1}, o_{t}, \mathcal{I}_t) $
            \STATE $ \mathcal{R}_{c_t}, \mathcal{R}_{s_t} \xleftarrow{} \textit{Explorer}(\text{Effect}(a_t), \mathcal{T}_i, \text{Objective}_i, o_{t+1}, \mathcal{H}_{t}) $
            \STATE $S_{\text{eff}}(a_t), S_{\text{fut}}(o_{t+1}) \xleftarrow{} \textit{Appraiser}(\text{Effect}(a_t), o_{t+1}, \mathcal{T}_i)$ \hfill $\triangleright$ \textbf{DES}
            \STATE $\mathcal{S}_{\text{total}}(a_t, o_{t+1}) \xleftarrow{} w_{\text{eff}} \cdot S_{\text{eff}}(a_t) + w_{\text{fut}} \cdot S_{\text{fut}}(o_{t+1}) $
            \STATE $ Q(s_{t+1}) \xleftarrow{} \mathcal{S}_{\text{total}}(a_t, o_{t+1}) $
            \STATE $ \mathcal{C}_t \xleftarrow{} \textit{Controller}(\mathcal{T}_i, \mathcal{H}_{t+1}, o_{t+1}) $
            \IF{ $ \mathcal{C}_t $ indicates stop}
	            \STATE break
            \ENDIF
            \STATE $ a_{t+1_{\text{sim}}}, \mathcal{I}_{t+1_{\text{sim}}} \xleftarrow{} \textit{Explorer}(o_{t+1}, \mathcal{T}_i, \mathcal{H}_{t+1}, \mathcal{C}_t) $ 
            \STATE $ s_{t+1_{\text{sim}}}, o_{t+1_{\text{sim}}} \xleftarrow{} \mathcal{F}(a_{t+1_{\text{sim}}}, s_{t+1}) $ 
				\STATE $ \text{Effect}(a_{t+1_{\text{sim}}}) \xleftarrow{} \textit{Explorer}(o_{t+1_{\text{sim}}}, o_{t+1}, \mathcal{I}_{t+1_{\text{sim}}}) $
            \STATE $ \mathcal{R}_{\text{sim}} \xleftarrow{} \textit{Explorer}(\text{Effect}(a_{t+1_{\text{sim}}}), \mathcal{T}_i, o_{t+1_{\text{sim}}}, [\mathcal{H}_{t+1}, a_{t+1_{\text{sim}}}]) $
        \FOR{ $ t' \xleftarrow{} t,\ldots,0 $ }
                \STATE $ Q(s_{t'}) \xleftarrow{} \max (Q(s_{t'+1}), Q(s_{t'}) ) $ \hfill $\triangleright$ \textbf{MVB}
        \ENDFOR
        \ENDWHILE         
    \ENDIF
    \STATE $ \text{Comp}_t, \mathcal{R}_{\text{sub}} \xleftarrow{} \textit{Controller}(\mathcal{T}_i, \mathcal{H}_{t+1}, o_{t+1}) $ \hfill $\triangleright$ \textbf{RTA}
    \IF{$ \text{Comp}_t $ indicates complete}
        \STATE $ s_0 \xleftarrow{} s_{t+1}, o_0 \xleftarrow{} o_{t+1} $ 
        \STATE $ \mathcal{P}' \xleftarrow{} [\mathcal{P}', \mathcal{T}_i]$
    \ENDIF
    \STATE $ \mathcal{P} \xleftarrow{} \textit{Planner}(\mathcal{P}, \text{Comp}_t) $
\ENDWHILE
\end{algorithmic}
\end{algorithm}

\end{document}